\documentclass[journal]{IEEEtran}
\usepackage[lined,boxed,commentsnumbered,ruled]{algorithm2e} 
\usepackage{hyperref}
\usepackage{multirow}
\usepackage{amsmath}
\usepackage{amssymb}
\usepackage{colortbl}
\usepackage{dashrule}
\usepackage{booktabs}
\usepackage{subfigure} 
\usepackage{graphicx}
\usepackage{harpoon}
\usepackage{epstopdf}
\usepackage{CJKutf8}

\usepackage{threeparttable}  
\usepackage{multirow,booktabs,color,soul,threeparttable}
\definecolor{hl}{rgb}{0.75,0.75,0.75}
\sethlcolor{hl}
\usepackage{caption}

\usepackage{rotating}
\usepackage[table*]{xcolor}
\DeclareGraphicsExtensions{.pdf,.png,.jpg,.eps}
\usepackage{bbding}
\usepackage{pifont}

\begin{document}
\begin{CJK}{UTF8}{gbsn}
\title{\textcolor{black}{Surrogate-Assisted Evolutionary Reinforcement Learning 
Based on Autoencoder and Hyperbolic Neural Network}
% 提升的话，可以用GAN或者Autoencoder或者VAE来降维；surrogate可以HNN或者DNN或者GNN。
}
\author{ 
 Bingdong~Li$^{1,2,3,4}$,
 Mei~Jiang$^{1,2,3}$,
 Hong~Qian$^{1,2,3}$,
 Ke~Tang$^{5}$,~\IEEEmembership{Fellow,~IEEE,}
 Aimin~Zhou$^{1,2,3}$,~\IEEEmembership{Senior~Member,~IEEE,}
 Peng~Yang$^{5,6}$,~\IEEEmembership{Senior~Member,~IEEE}\thanks{Peng~Yang is the corresponding author.}\\
 $^{1}$Shanghai Institute of AI for Education, East China Normal University, Shanghai 200062, China\\
$^{2}$School of Computer Science and Technology, East China Normal University, Shanghai 200062, China\\
$^{3}$Shanghai Frontiers Science Center of Molecule Intelligent Syntheses\\
$^{4}$Key Laboratory of Advanced Theory and Application in Statistics and Data Science, Ministry of Education, China\\
$^{5}$The Guangdong Key Laboratory of Brain Inspired Intelligent Computation, Department of Computer Science and Engineering, Southern University of Science and Technology, Shenzhen 518055, China\\
$^{6}$Department of Statistics and Data Science, Southern University of Science and Technology, Shenzhen 518055, China\\
\texttt{bdli@cs.ecnu.edu.cn, 51265901071@stu.ecnu.edu.cn,}\\
\texttt{hqian@cs.ecnu.edu.cn, tangk3@sustech.edu.cn,}\\
\texttt{amzhou@cs.ecnu.edu.cn, yangp@sustech.edu.cn}
 %Hong~Qian,
 %Ke~Tang,~\IEEEmembership{Fellow,~IEEE,} 
% %Jian~Jin
% and
% Aimin~Zhou,~\IEEEmembership{Senior~Member,~IEEE}
\thanks{ 
% Corresponding author: A. Zhou.
Manuscript received –.
% This work was supported
% by the Science and Technology Commission of Shanghai Municipality Grant (No. 22511105901),
% the Fundamental Research Funds for the Central Universities,
% in part by National Natural Science Foundation of China (Grants 62272210 and 62250710682),
% in part by the Program for Guangdong Introducing Innovative and Entrepreneurial Teams (Grant 2017ZT07X386), 
% in part by the Guangdong Provincial Key Laboratory under Grant 2020B121201001.
% \textit{(Corresponding author: Peng Yang)}

% B. Li, M. Jiang, and A. Zhou are with the
% Lab of Artificial Intelligence for Education, East China Normal University, 3663 Zhongshan North Road, Shanghai, 200062, China.
% They are also with the
% School of Computer Science and Technology, East China Normal University, 3663 Zhongshan North Road, Shanghai, 200062, China  
% (e-mails:
% bdli@cs.ecnu.edu.cn;
% 51265901071@stu.ecnu.edu.cn;
% % 51255901098@stu.ecnu.edu.cn;
% amzhou@cs.ecnu.edu.cn).

% P. Yang is with the Research Institute of Trustworthy
% Autonomous Systems, and Guangdong Key Laboratory of Brain-Inspired
% Intelligent Computation, Department of Computer Science and Engineering,
% Southern University of Science and Technology, Shenzhen 518055, China. 
% P. Yang is also with the Department of Statistics and Data Science, Southern University of Science and Technology, Shenzhen 518055, China.
% (e-mail: yangp@sustech.edu.cn). %; xiny@sustech.edu.cn).
}

% \thanks{
% Copyright (c) 2012 IEEE. Personal use of this material is permitted. However, permission to use this material for any other purposes must be obtained from the IEEE by sending a request to pubs-permissions@ieee.org.
% } 
 
    }
    
%\author{IEE222222E Publication Technology,~\IEEEmembership{Staff,~IEEE,}
%        % <-this % stops a space
%\thanks{This paper was produced by the IEEE Publication Technology Group. They are in Piscataway, NJ.}% <-this % stops a space
%\thanks{Manuscript received April 19, 2021; revised August 16, 2021.}}

\markboth{Journal of \LaTeX\ Class Files,~Vol.~14, No.~8, August~2021}%
{Shell \MakeLowercase{\textit{et al.}}: A Sample Article Using IEEEtran.cls for IEEE Journals}


\maketitle

\begin{abstract}
Evolutionary Reinforcement Learning (ERL), training the Reinforcement Learning (RL) policies with Evolutionary Algorithms (EAs), has demonstrated enhanced exploration capabilities and greater robustness than using traditional policy gradient. However, ERL suffers from the high computational costs and low search efficiency, as EAs require evaluating numerous candidate policies with expensive simulations, many of which are ineffective and do not contribute meaningfully to the training. One intuitive way to reduce the ineffective evaluations is to adopt the surrogates. Unfortunately, existing ERL policies are often modeled as deep neural networks (DNNs) and thus naturally represented as high-dimensional vectors containing millions of weights, which makes the building of effective surrogates for ERL policies extremely challenging. 
This paper proposes a novel surrogate-assisted ERL that integrates Autoencoders (AE) and Hyperbolic Neural Networks (HNN). Specifically, AE compresses high-dimensional policies into low-dimensional representations while extracting key features as the inputs for the surrogate. HNN, functioning as a classification-based surrogate model, can learn complex nonlinear relationships from sampled data and enable more accurate pre-selection of the sampled policies without real evaluations. 
The experiments on 10 Atari and 4 Mujoco games have verified that the proposed method outperforms previous approaches significantly. The search trajectories guided by AE and HNN are also visually demonstrated to be more effective, in terms of both exploration and convergency. This paper not only presents the first learnable policy embedding and surrogate-modeling modules for high-dimensional ERL policies, but also empirically reveals when and why they can be successful.
%the superiority of the  over s, including
%problem transformation based algorithm, decision variable
%clustering based algorithm, particle swarm optimization algorithm,
%and estimation of distribution algorithm.

%
%?????????????????????????????????????????
%???????????????????????? (model-free) Fitness Aware Search Operator (FASO) ????????(????????IBEA ??SMS-EMOA??)
%?????????convergence subpopulation?CP??diversity subpopulation (DP)??archive?
%??????????Fitness Aware Search Operator (FASO) ?DP?CP??????????
%????????????reference vector?adaptive PBI????????????????????????????????0.1?Theta?5

%problem, competitive
%swarm optimizer, particle swarm optimization
%
%Adaptive offspring generation, 
%large-scale, multiobjective optimization.
%
%Terms?Large-scale Optimization, Multiobjective Op-timization, Differential Evolution, Variable Grouping.

%This document describes the most common article elements and how to use the IEEEtran class with \LaTeX \ to produce files that are suitable for submission to the IEEE.  IEEEtran can produce conference, journal, and technical note (correspondence) papers with a suitable choice of class options. 
\end{abstract}

\section{Introduction}
Evolutionary Reinforcement Learning (ERL), an emerging technique that combines Evolutionary Algorithms (EAs)\cite{back1993evooverview,qian2025securitygames} and Reinforcement Learning (RL)\cite{such2017neuroevolution}, has garnered widespread attention, particularly with the rise of Deep Neural Networks (DNNs) in recent years. ERL leverages the global search capabilities of EAs and the Markov Decision Process modeling of RL to address sequential decision-making problems in complex environments~\cite{bai2023evolutionary}. Compared to traditional RL methods, ERL exhibits superior exploration capabilities, higher noise resistance, and ease of parallelization, making it particularly effective for high-dimensional and complex tasks~\cite{sigaud2023survey}. This approach has been successfully applied to various domains, including video games~\cite{salimans2017evolution}, scheduling in cloud computing~\cite{yang2024reducing}, robotic control~\cite{yuan2022euclid}, and supply chain management~\cite{ni2021multi}.

However, it is important to distinguish between the different algorithmic paradigms that fall under the umbrella of evolutionary policy search. According to the recent taxonomy proposed by S. O~\cite{sigaud2023survey}, such methods can be classified into three categories: (1) \textbf{EA-based policy optimization}, where policies are optimized purely through EAs without any RL component; (2) \textbf{Evolutionary Reinforcement Learning}, where EAs and RL components interact cooperatively during training; and (3) \textbf{RL-dominated methods with EA modules}, in which RL serves as the core optimizer and EAs are used for enhancing structural or exploration diversity.

Our method clearly belongs to the first category—EA-based policy optimization. It directly evolves DNN-based policies via a black-box evolutionary algorithm (NCS)\cite{tang2016ncs} and does not involve gradient-based updates, value functions, or actor-critic structures. Thus, it is more accurately described as an EA-based method rather than ERL in the hybrid sense.

%In the ERL framework, EAs and RL play distinct yet cooperative roles in policy optimization, with EAs responsible for policies generation and RL for policies evaluations\cite{salimans2017evolution}. EAs explore the policy space by generating candidate policies through randomized search operators like mutation and crossover, while RL interacts with the environment step-by-step to rollout and evaluate these policies' performance and provides reward signals as fitness metrics for EAs. Through selection mechanisms, EAs retain highly fitted policies and eliminate less fitted ones, iteratively optimizing the policy. This combination allows EAs' global search capabilities to complement traditional RL's shortcomings in exploration. However, despite its unique advantages\cite{Sigaud2023combin,hao2022ERL-RE}, ERL still faces two core challenges: low search efficiency and high computational costs.

Despite their different formulations, a common challenge across all categories of ERL—including hybrid ERL and EA-based policy search—is the high computational cost associated with evaluating large numbers of candidate policies in complex environments. This issue is particularly severe when policies are represented by deep neural networks with millions of parameters. Evaluating each policy requires costly environment rollouts, which limits scalability and sample efficiency across the entire family of ERL methods\cite{salimans2017evolution, chrabaszcz2018back}.

This difficulty arises from two main sources. First, as DNN-based policies involve millions of parameters, the search space expands exponentially, making it harder to explore effectively. Second, determining accurate search directions becomes more computationally expensive, as reliable policy evaluation through environment interactions is needed but hard to obtain due to the instability or high variance in value estimation. These twin challenges significantly impact both the effectiveness and the efficiency of the evolutionary search process\cite{wang2022scerl, stork2019neuroevolution}.
%To be specific, since DNNs have been preferred as RL policies, their weights serve as the parameters of the policy to be optimized and often come to the scale of millions, leading to an exponential growth of the search space in the EA phase. This largely increases not only the difficulty of effective exploration but also the computational costs\cite{Sigaud2023combin}. On one hand, as the search space enlarges exponentially, the samples required by EAs also increase significantly, due to its randomized iterative search nature. On the other hand, determining the search direction at each iteration becomes more computational costly, because the learning of the action selections needs more interactions with the environment, given the less accurate Q-value estimation. Both issues impede the adaptability of ERL methods and limit their applicability in complex real-world scenarios.
%--------------------------

%In EAs, the issues of low search efficiency and high computational costs can be ideally jointly addressed by the Surrogate-Assisted methods\cite{2024It}, which work by building computationally cheap models (i.e., surrogates) for approximating the expensive evaluations and adaptively replacing the real evaluations of the less promising solutions with the surrogates. In this regard, large volume of computational costs on the ineffective candidate solutions can be saved. Unfortunately, existing surrogate-assisted methods cannot be trivially incorporated into ERL. 
Surrogate-assisted approaches\cite{jin2011saec} have been widely adopted to alleviate this problem. These methods build inexpensive surrogate models to approximate the expensive evaluation process and reduce the number of real environment interactions. While promising in theory, their effectiveness in ERL remains limited in practice. Most existing surrogate models either assume low-dimensional optimization problems\cite{stork2019neuroevolution, francon2020effective} or apply to action-level evaluations such as critic networks\cite{Lillicrap2016DDPG, haarnoja2018sac}, which do not directly assess whole-policy performance. Moreover, in high-dimensional policy spaces, it is difficult to train accurate surrogates due to complex nonlinearities and sparse sample availability.

Although several ERL works have explored surrogate modeling in some form, most do not explicitly address the challenge of high-dimensional DNN policies. PE-SAERL\cite{tang2022pe} is among the few that directly tackle this issue by embedding policies into a lower-dimensional space using random projections and applying a fuzzy classifier as a surrogate. However, PE-SAERL uses fixed modules that lack adaptivity, and its effectiveness is constrained by embedding quality and surrogate accuracy.
Furthermore, PE-SAERL leaves two critical questions unanswered: (1) why can pre-selection in a reduced space still be effective, given the inevitable information loss from dimensionality reduction? and (2) if the reduced space is effective, why not conduct the entire EA-based search in that space?

%There are also several ERL methods that attempt to use surrogate models to replace real policy evaluations instead of immediate rewards\cite{stork2019surrogate,wang2022SCERL,stork2019improving,francon2020effective}. Nevertheless, most of them did not consider parametric DNN as the policy, thus they did not encounter the issue of high-dimensional parameters. 
%Recently, Tang et al.\cite{tang2022enabling} introduced the first framework explicitly dealing with the high-dimensional issue while incorporating surrogate-assisted module into ERL, called Policy-Embedding Surrogate-Assisted ERL (PE-SAERL). PE-SAERL first projects the policy into a lower-dimensional space with Random Embedding and then builds a Fuzzy Classification Pre-Selection (FCPS) surrogate in the reduced space. However, both the dimension reduction module and the surrogate-assisted module module in PE-SAERL lack the learning capabilities and thus result in unsatisfactory surrogate accuracy. Most importantly, PE-SAERL left the answers of two fundamental questions of the PE-SAERL framework blank: 1) why the pre-selection in low-dimensional space can be successful given the information loss of the dimension reduction; 2) if low-dimensional is beneficial, why not also conduct the EA search in low-dimensional space? 

Building upon the above discussions, this paper aims to propose a novel SAERL featured with learnable dimension reduction for the policy embedding and a learnable approximation for ranking pairwise candidate policies in the reduced space, leading to a much powerful SAERL method than PE-SAERL. Besides, this paper aims to answer the above two questions left by PE-SAERL for more in-depth understanding of policy embedding in ERL.
First, Autoencoder (AE) is integrated into the ERL framework to address the curse of dimensionality. Rather than feeding high-dimensional policy vectors into a surrogate model, an AE network is trained to map these vectors from high-dimensional to low-dimensional representations. Compared to the random embedding in PE-SAERL, AE offers the ability of better capturing essential data features and performing complex transformations using multiple hidden layers and nonlinear activation functions.
Hyperbolic Neural Network (HNN) is then adopted as the surrogate model to classify candidate solutions into favorable and unfavorable groups. The most promising solutions among the favorable ones are chosen as the offspring for the EAs. By evaluating only one candidate solution from the most promising ones during each iteration, this approach significantly reduces time costs.
%在每次迭代中，仅从挑选出的候选解中评估一个解（或少量解），从而减少了真实环境的评估次数，降低了整体的计算成本。
Compared to traditional classification models, HNN embeds the inputs into the hyperbolic space, enhancing its ability to capture nonlinear relationships and data features, thereby improving the prediction accuracy.
Following PE-SAERL, NCS is also employed as the EA-based trainer, which has been mathematically shown to be suitable for multi-modal optimization problems\cite{yang2021parallel}.
Moreover, compared with Wang et al.\cite{wang2022scerl} who use a VAE-GP surrogate in a two-stage training pipeline, our method introduces a jointly learnable AE-HNN surrogate. It supports end-to-end optimization, leverages hyperbolic geometry for better structure preservation, and enables soft-ranking for policy filtering. Unlike GP-based regression which suffers from scalability issues in high-dimensional policy spaces, our HNN surrogate performs robustly even on millions of parameters.
Extensive empirical studies on 10 Atari and 4 Mujoco tasks have been conducted against compared algorithms. The main contributions of this work are as follows:
 \begin{enumerate}
 \item \textbf{The first LEARNABLE dimensionality reduction and surrogate modeling method for tackling the challenges of training high-dimensional ERL policies}, enhancing the ability of capturing complex features of the high-dimensional policy space and leading to much better EA-based policy search.
 %\item \textcolor{black}{Search-Driven Dimensionality Reduction and pre-selection: A novel "search-dimensionality reduction-pre-selection" framework is proposed to address the challenges of optimizing high-dimensional policy spaces. Experiments demonstrate that while dimensionality reduction inevitably introduces information loss, low-dimensional searches are not equivalent to high-dimensional searches, but low-dimensional rankings are close to high-dimensional rankings.}
  \item \textbf{The condition of successfully operating in the heavily reduced space is empirically revealed}, given the inevitable information loss of dimensionality reduction. Briefly, it works when the ranks of the reduced low-dimensional population keep highly consistent with its original ranks in the high-dimensional space. This finding makes the SAERL framework more convincing and tractable.
 \item We empirically show how the proposed method of \textbf{pre-selecting in low-dimensional space can lead to more diverse exploration and guided convergence}, which is expected to benefit the broader ERL methods.
 %\item \textcolor{black}{Enhanced Exploration and Convergence: Experiments on Atari and Mujoco tasks demonstrate that the proposed method not only reduces computational costs but also significantly improves exploration efficiency and convergence speed in evolutionary reinforcement learning.}
 \end{enumerate}
 
 The remainder of this paper is structured as follows:
Section \ref{Background} introduces the related works and preliminary background of 
this work;
  Section \ref{sectionOurApproach} presents the details of the proposed algorithm; 
%  Section \ref{sectionOurApproachArchive} illustrates SRA with archive;
  Section \ref{sectionExperimentalSetup} is devoted to experimental setup.
  Section \ref{sectionExperimentalResults} discusses the experimental results,  
  Section \ref{sectionConclusion} concludes the paper and indicates some future directions.

\section{Background}\label{Background}
\subsection{Evolutionary Reinforcement Learning}

\begin{figure*}[t]   
   \centering
     \includegraphics[width=400pt]{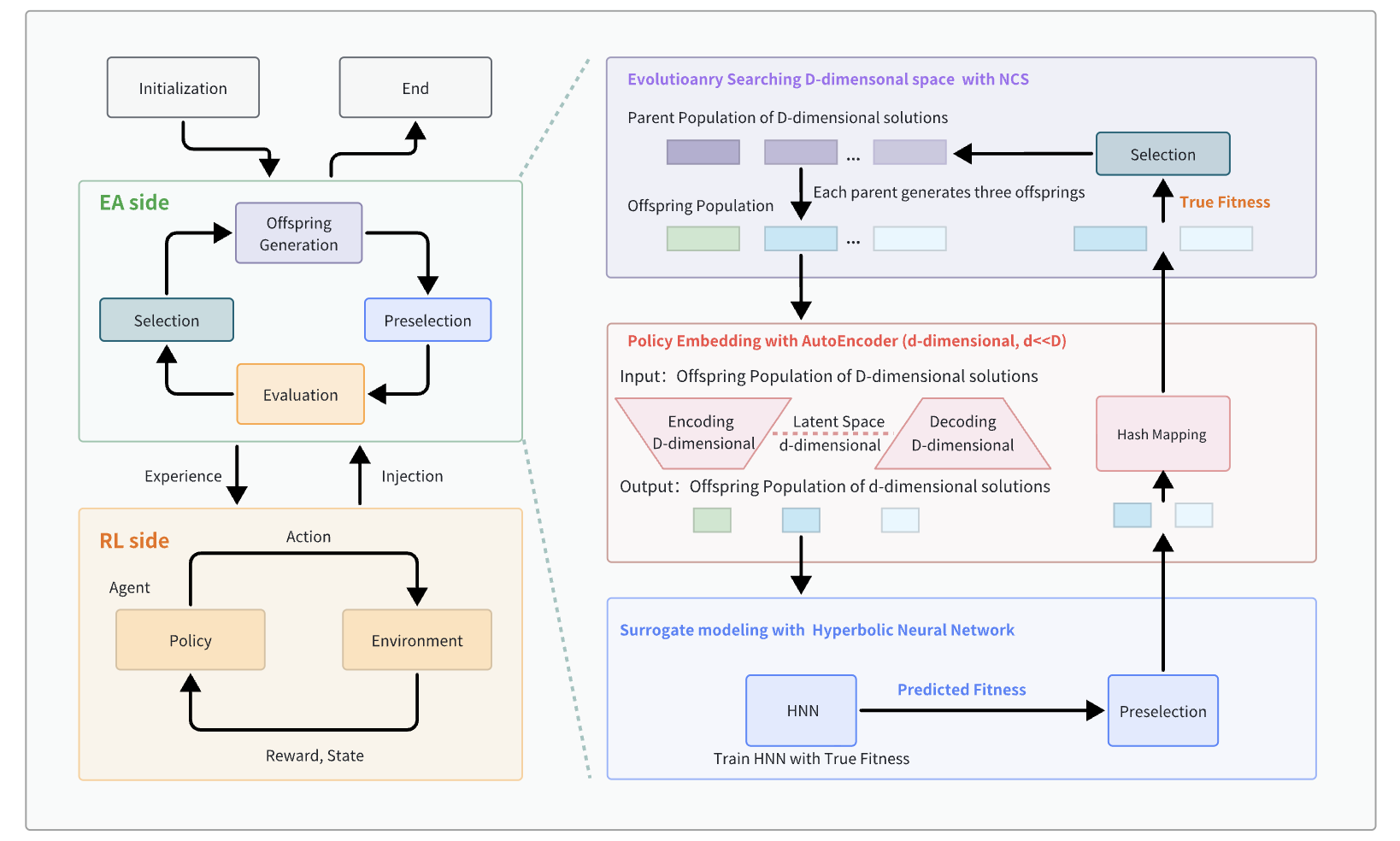}  %%%lbd20150622 %%%lbd20160114
   \caption{\textcolor{black}{The proposed ERL framework integrates an Autoencoder and a Hyperbolic Neural Network for enhanced dimensionality reduction and policy pre-selection. The left part shows the traditional ERL workflow with EA and RL phases. Our algorithm innovates in the EA phase. We first use an Autoencoder to embed offspring policies to a lower-dimensional space. Then, the Hyperbolic Neural Network acts as a surrogate model, predicting the rank of policies' qualities. This enables efficient pre-selections, reducing unnecessary evaluations. After pre-selection, the selected policies are mapped to the high-dimensional space for real fitness evaluations, and then are incorporated into the evolutionary search process. }
   }
   
   \label{AEHNNNCS}
\end{figure*}
The integration of Evolutionary Algorithms (EAs) and Reinforcement Learning (RL) has led to the development of a broad class of algorithms aimed at improving policy optimization in sequential decision-making. According to the taxonomy proposed by S. O ~\cite{sigaud2023survey}, these methods can be categorized into three main types:
 \begin{enumerate}
 \item \textbf{EA-based policy optimization}, also known as \textit{Neuroevolution}, in which candidate policies—typically parameterized as neural networks—are optimized purely through evolutionary search without using RL techniques such as temporal difference learning or policy gradients. Representative methods include Genetic Algorithms (GAs), Natural Evolution Strategies (NES), and OpenAI's Evolution Strategies (OpenES)~\cite{salimans2017evolution, chrabaszcz2018back}.
\item \textbf{Evolutionary Reinforcement Learning (ERL)}, where EAs and RL modules operate in parallel or collaboratively. In these methods, EAs often generate diverse policies, while RL improves them via actor-critic or Q-learning components. Examples include RE$^2$-ERL~\cite{hao2022erlre2} and CERL~\cite{khadka2019cerl}.
\item \textbf{RL-dominated methods with EA modules}, which mainly rely on gradient-based RL for policy optimization but integrate EA components for structural search or diversity enhancement, such as G2PPO and SUPE-RL~\cite{Marchesini2021SUPE}.
 \end{enumerate}
 
Our method falls into the first category, i.e., EA-based policy optimization. It uses NCS to directly evolve policy networks without any critic network or policy gradient mechanism. This positions our work as a gradient-free alternative to ERL methods.

\subsection{Surrogate-Assisted Evolutionary Algorithms}
The Surrogate-Assisted Evolutionary Algorithm (SAEA) is a method that integrates EAs with surrogate models to address the inefficiencies of EAs in solving computationally expensive problems (CEPs). By approximating the fitness landscape, SAEA reduces the need for costly fitness evaluations and improves convergence speed and global search capabilities.

SAEAs generally employ surrogate models to guide evolutionary searches. \textcolor{black}{For instance, multi-objective optimization has been leveraged to integrate the capabilities of multiple large language models (LLMs), demonstrating enhanced adaptability and performance in tasks requiring diverse expertise\cite{li2024llmmerge}.} Based on how they predict fitness values, surrogate approaches are categorized into absolute fitness models and relative fitness models:

\begin{enumerate}
    \item Absolute Fitness Models:
These models approximate the true fitness function directly. Common regression-based techniques include Polynomial Regression (PR), Support Vector Regression (SVR), Kriging, and Radial Basis Functions (RBF)\cite{Forrester2009Recent}. Kriging is especially popular for its ability to estimate uncertainty alongside fitness values, while SVR has been applied to practical problems like wind barrier optimization\cite{llora2005combating}. Simpler similarity-based models, such as k-nearest neighbor regression (kNN-R), are computationally cheaper, focusing solely on the similarity between individuals.
    \item Relative Fitness Models:
Instead of predicting absolute values, these models rank or classify candidate solutions within a population. For instance, classification-assisted methods like CADE\cite{lu2011classification} use classifiers (e.g., SVC or fuzzy-kNN) to filter unpromising candidates before evaluation. However, these models focus on local fitness trends and lack the ability to provide global insights into the search space.
\end{enumerate}

SAEAs have demonstrated great potential in solving CEPs, yet their integration into ERL remains underexplored. This presents an opportunity to investigate how surrogate models can enhance ERL, particularly in addressing challenges related to high-dimensional policy spaces and computational efficiency.

\subsection{Surrogate Modeling in Evolutionary Reinforcement Learning}
To reduce the high computational cost of real-environment evaluations, surrogate-assisted approaches have been adopted in both traditional EAs and ERL~\cite{jin2011saec}. These approaches use surrogate models to approximate the reward or fitness of candidate solutions, thereby filtering out unpromising candidates before actual evaluation.
There are generally two types of surrogate models used in this context:
 \begin{enumerate}
\item \textbf{Action-level surrogates}, such as critic networks in RL (e.g., DDPG~\cite{Lillicrap2016DDPG}, SAC~\cite{haarnoja2018sac}), which approximate the value function but only operate on single states or actions. These methods do not model the overall performance of a policy trajectory and hence are limited in guiding full policy selection in EA.

\item \textbf{Policy-level surrogates}, which aim to approximate the performance of an entire policy. Existing efforts include using regression models~\cite{wang2022scerl} or classification-based selection schemes~\cite{stork2019neuroevolution, francon2020effective}. However, these works mostly operate in low-dimensional or simplified control settings.
\end{enumerate}
The major challenges in applying surrogate models in high-dimensional ERL include:
\begin{enumerate}
\item The difficulty of learning accurate mappings from millions of policy parameters to scalar fitness values.
\item The nonlinearity and sparsity of policy space, which degrade surrogate reliability.
\item The limited generalization capacity of fixed surrogate architectures.
\end{enumerate}
PE-SAERL~\cite{tang2022pe} is a notable exception that tries to address these issues by introducing random projection for dimensionality reduction and applying a fuzzy classification surrogate. However, it does not learn the projection or surrogate components adaptively, limiting its flexibility and precision.

To overcome these limitations, our proposed framework introduces learnable components: an Autoencoder (AE) for adaptive low-dimensional representation learning, and a Hyperbolic Neural Network (HNN) for classification-based surrogate modeling. These enhancements improve both expressiveness and ranking accuracy, facilitating more efficient policy pre-selection.
\section{The Proposed Approach}\label{sectionOurApproach}
% \textcolor{black}{Overview}

% \textcolor{black}{AE}

% \textcolor{black}{HNN}
This section first discusses the core questions in the framework design of the proposed approach, named AE-HNN-NCS. And then two key components of AE-HNN-NCS are detailed.

\subsection{The Framework}

This framework of AE-HNN-NCS is depicted in Fig.\ref{AEHNNNCS}. The core motivation behind our proposed method lies in addressing two fundamental limitations observed in prior surrogate-assisted ERL approaches: the lack of adaptive representation for high-dimensional policy parameters, and the limited modeling capacity of conventional surrogates when handling complex, nonlinear decision boundaries.

\textbf{Autoencoder (AE) as a Learnable Embedding Module.} Traditional surrogate approaches often rely on fixed or random embeddings (e.g., random projection in PE-SAERL) to reduce the dimensionality of policy representations. However, such embeddings cannot adapt to the underlying structure of the data and are prone to discarding task-relevant features. In contrast, AEs provide a data-driven approach to learning compact, informative latent spaces by minimizing reconstruction loss. This allows the policy representations in the embedded space to preserve critical semantic information, improving the reliability of downstream surrogate predictions.

\textbf{Hyperbolic Neural Network (HNN) as a Surrogate.} Classical surrogate models such as support vector classifiers or fuzzy k-NN often operate in Euclidean space and lack the expressiveness needed for modeling complex distributions in policy space. HNNs extend conventional neural networks into hyperbolic geometry, which is better suited for representing hierarchical, tree-like, or strongly clustered structures. This geometric flexibility enables HNNs to better capture subtle distinctions between promising and unpromising policies, leading to more accurate pre-selection and enhanced sample efficiency.

By combining AE and HNN, our framework forms a fully learnable surrogate-assisted policy optimization mechanism. The AE compresses policies into structured low-dimensional features, and the HNN leverages this embedding to perform reliable binary classification for policy filtering. This design offers greater adaptability and generalization compared to prior static or low-capacity surrogates.%Following the general ERL framework, in AE-HNN-NCS, EA works for iteratively sampling the individuals, each of which represents the weight parameters of a DNN-based policy. And RL works for rollouting each policy in the environment and returning a reward for evaluating the policy. Following PE-SAERL, the NCS algorithm is employed as the EA-based trainer in the original policy space, and a policy embedding module and a surrogate-assisted module are also adopted. Here, in each search iteration of NCS, each candidate individual (i.e., the weights of a DNN-based policy) is first embedded with policy embedding module into low-dimensional, and then pre-selected by the surrogate-assisted module based on the low-dimensional representation. After that, the hash mapping copies of the pre-selected individuals in the original high-dimensional space are evaluated with real objective functions and then used for generating the offspring individuals of the next iteration by NCS. 

%What is different from PE-SAERL is that the policy embedding module and the surrogate-assisted module in AE-HNN-NCS is learnable, i.e., AE for replacing the random embedding of PE-SAERL, and HNN for substituting the FCPS module of PE-SAERL. With the learnability introduced by AE and HNN, the proposed AE-HNN-NCS is able to adaptively capture the complex characteristics of different environments and thus leads to much higher surrogate accuracy and ERL training performance.

%Before detailing the AE-based policy embedding module and the HNN-based surrogate-assisted module, this paper first discusses why we (and PE-SAERL) only conduct the pre-selection in the embedded low-dimensional space, while leaving the EA-based search in the original space. In other words, why the embedded low-dimensional policies are beneficial for the pre-selection, yet they are ineffective for searching? These two important questions serve as the foundation of the policy-embedding idea for surrogate-assisted ERL, but have not been touched by PE-SAERL in its paper\cite{tang2022enabling}. 

\subsection{Why Pre-select, not Reproduce, in the Embedded Space?}
From the high-dimensional original policy space to the low-dimensional embedded policy space, it is highly likely that certain information may be lost. This means that the landscape of the two space can be different and thus the distribution of fitness may be changed to some extent during the dimension reduction. This suggests that any operators that utilize the accurate fitness values may fail within the low-dimensional space. 
On the other hand, as long as the changes of the fitness distribution do not re-order the fitness rank of the policies, any operators that only utilize the fitness ranks can be still useful in the low-dimensional space. 

For example, the crossover operator of differential evolution and the roulette selection operator of genetic algorithm are based on the exact fitness values. Hence, search in low-dimensional space may result in different offspring distribution from searching in the original high-dimensional space. If so, search in the low-dimensional space can be ineffective or even harmful to the surrogate-assisted ERL.
Some operators like $(\lambda,\mu)$ selection in evolution strategies only require the information of fitness ranks of the population. If the embedded low-dimensional space preserves the same fitness ranks with the high-dimensional space, even in local regions, the policy embedding idea can work. To summarize, keep the fitness ranks similar between high-dimensional space and low-dimensional space similar is much simpler than keeping the exact fitness values similar between the two space.

This suggests that the surrogate assisted module in our framework should adopt the pre-selection based surrogates, rather than the fitness regression based surrogates. Similarly, many reproduction operators require the exact fitness values of the parent population, therefore we suggest conducting searching in the original space rather than the embedded space. Section \ref{whypre-select} provides detailed empirical verification for the above discussions.

This also relies an important assumption that in RL environment, we can have a powerful policy embedding strategy that successfully keeps at least the fitness rank between high-dimensional space and low-dimensional space similar. That is why we propose the learnable AE rather than using the original randome embedding. 

\begin{figure}[ht]   
   \centering
     \includegraphics[width=200pt]{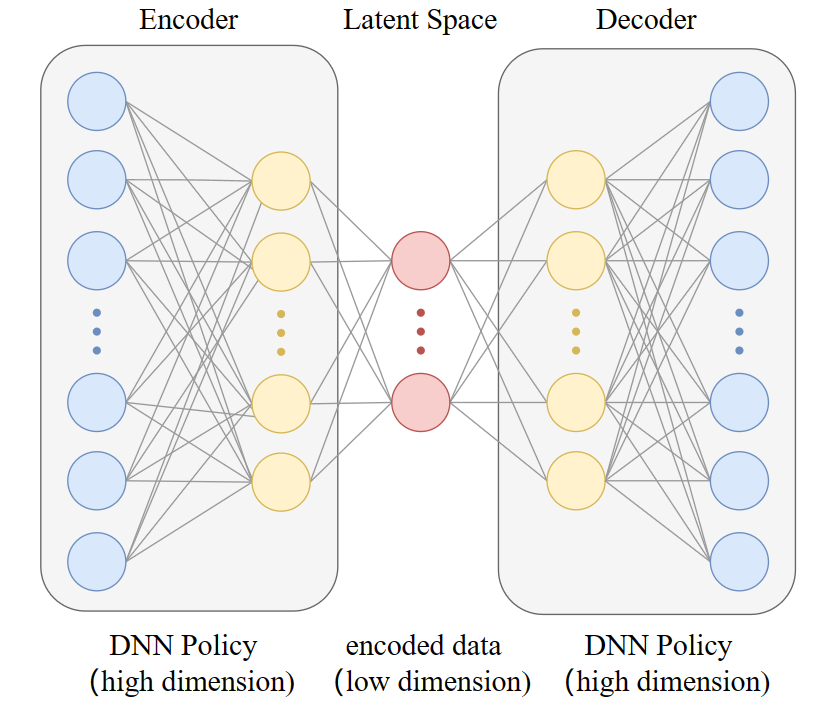}  %%%lbd20150622 %%%lbd20160114
   \caption{\textcolor{black}{The overview of autoencoder.}
   % \textcolor{black}{The text font size
   % in this figure 
   % is too small.
   % you need to edit the original figure file instead of the width in latex
   % to make the font size right.\\
   % }
% \textcolor{black}{font size of lower texts still not good ;
% Information density too low;
%  ask yyt how to draw figures with appropriate font size.}
   }
   \label{AE}
\end{figure}
\subsection{The AE-based Policy Embedding Module}
The autoencoder, shown in the Fig.\ref{AE}, which is an unsupervised learning method, has become increasingly popular in representation learning\cite{hinton2006reducing,masci2011stacked}. It can reduce classifier training iterations and boost accuracy\cite{erhan2010pretraining,le2013unsupervised}. This is mainly due to two aspects: it selects and represents key features better than the original input, and its encoder maps high-dimensional inputs to a low-dimensional latent space, enhancing subsequent model efficiency and reducing overfitting risk.
On this basis, we propose to use AE for representing DNN-based RL policies into low-dimensional. 

Another underlying assumption is that, the true rewards of the RL policy are only influenced by a subset of the features, rather than all of them. This assumption is technically justifiable for DNN-based policies. DNN models often have redundant weights from the convolutional layer to the fully-connected layer, and a large number of neuron activations tend to approach zero. Even after removing these neurons, the model can still maintain the same expressive power, a phenomenon known as over-parameterization, with the corresponding technique called model pruning\cite{li2022stageprune}.

Additionally, AE does not require class labels for data during its establishing. Even though it introduces extra steps for pretraining the AE model, the cost is much lower compared to the overall efficiency improvement brought by AE. This is because AE's training data does not need reinforcement learning rewards, meaning we can obtain this data from the environment around the randomly acting agents or robots without any specific operations. Therefore, using AE can potentially save computational resources and reduce the overall training cost in the reinforcement learning process.

The module includes the following steps: (1) Randomly generating policy samples and collecting them. (2) Training the autoencoder with the collected data. (3) Using the encoder of the trained autoencoder to perform dimensionality reduction on the policy samples.

Next, it is explained how to train the autoencoder. The autoencoder aims to minimize the difference or loss between the original data (the input of the encoder) and the reconstructed data (the output of the decoder). By minimizing the loss, the major information of the original data is considered to be represented by the low-dimensional latent vector (see Fig.\ref{AE} for illustration). This loss function is commonly referred to as the reconstruction error. Since our input is real-valued, mean squared error (MSE) is chosen as the loss function. Furthermore, we use ADAM as the optimizer. The two networks are initialized randomly and then iteratively optimized with the well-established ADAM optimizer by minimizing the loss. Mathematically, let the encoder and the decoder denote as $\phi$ and $\varphi$, respectively. And let the original vector (individual) as $\textbf{x} \in \mathrm{R}^n$, and the latent vector $\textbf{y} \in \mathrm{R}^m$, $n >> m$. The encoder compresses the input data into a low-dimensional latent representation, while the decoder reconstructs the input data from the latent representation. Let $\phi^*$ and $\varphi^*$ be the optimal encoder and decoder trained by the above descriptions, we have:
%向量用xy加粗
\begin{eqnarray}
%I_{SDE}(\textbf{x}, \textbf{y}) = \label{SDEcomputation}\\
&\phi : x \to y, \varphi : y \to x\\
&\phi^*, \varphi^* = \arg\min_{\textbf{$\phi$},\textbf{$\varphi$}}\mathbb{E}_{x \sim P(x)}\| x-( \varphi(\phi(x)))\|^2
\
\label{I2computation}
\end{eqnarray}

\textcolor{black}{An important aspect of our approach only adopt the encoder of the trained autoencoder. Such uses are typically seen in unsupervised learning tasks, such as feature extraction or dimensionality reduction. In our method, policy search is conducted in the high-dimensional space, and the autoencoder is used to represent the input feature vectors to address the curse of dimensionality, thereby improving the accuracy of the subsequent surrogate-assisted model. If the search is performed in the low-dimensional space, the reconstruction error introduced by the decoder part would lead to significant discrepancies when mapping individuals back to the original high-dimensional space. These discrepancies would lead to generated policies not accurately representing the originally selected ones. Therefore, in our method, to avoid reconstruction errors between the decoder output and the original input, we do not use the decoder. Instead, after the AE+HNN completes the selection in the low-dimensional space, the corresponding original high-dimensional policy is directly retrieved via indexing, ensuring that the final evaluation in the environment is conducted on the actual individual. This guarantees the overall consistency, reproducibility, and distortion-free process.}
%Instead, we pre-store the original input in an array and use hash mapping to directly map the low-dimensional representation back to the original input.}

\begin{algorithm} 
    %\footnotesize 
    \LinesNumbered
    %\SetAlgoNoLine
    \caption{HNN-Based Preselection} \label{HNNBasedpre-selection}   
    %\SetKwData{localTimerOfGroups}{localTimerOfGroups}
    %\SetKwData{globalTimer}{globalTimer}
    \SetKwData{Q}{Q}
    %\SetKwData{y}{y}
    \SetKwInOut{Input}{input}\SetKwInOut{Output}{output}
    \Input{Local encoded best policies \textbf{$y_i$}, \\
 Local best fit $f_i$, \\
 Candidate policies $y'_{i,1},y'_{i,2}...,y'_{i,M}$, \\
 }
    \Output{best candidate policy $y^*$}
    \BlankLine  
    $Labels$ $\gets$ PreparedLabels($x_i,f_i$); \\
    $Model$ $\gets$ TrainHNN($x_i,Labels$); \\
    Feed $y'_{i,1},y'_{i,2}...,y'_{i,M}$ into $Model$ to get prediction results $labels$; \\
    $indexs$ $\gets$ sort($labels$); \\
    $y^*$ = $indexs[0]$; \\
    return $y^*$;
%       Copy top $|U_t|/2$ solutions of $U_t$ to $P_{t+1}$
    \end{algorithm}

\subsection{The HNN-based Surrogate-Assisted Module}
%HNN作为preliminary，放到方法章节的HNN小节里，就像现在对AE的安排一样
In high-dimensional policy representation, traditional Euclidean space embedding methods often fail to capture complex nonlinear relationships effectively. In contrast, HNN leverage their unique geometric structure, which is more suitable for modeling hierarchical data and complex relationships in high-dimensional spaces\cite{nickel2017poincare, sala2018representation,gromov1987hyperbolic}. Thus, HNN is chosen as the surrogate model for policy pre-selection in this study, aimed at improving the efficiency of policy evaluation\cite{tifrea2019poincareglove, klimovskaia2020poincaremaps, khrulkov2020hyperimage}.

Given the policies are represented as low-dimensional vectors by AE, we discuss how to utilize an HNN as the surrogate model, which can be divided into three steps: a) Data preparation and processing, 
b) Training the agent model using HNN, and 
c) Using the agent model for pre-selecting RL policies.

% \paragraph{Data preparation}
Before training the HNN, we need to first prepare sufficiently reliable data and process the data. 
To reduce computational cost, we use an AE to map high-dimensional policy representations into a lower-dimensional space. %First, policy samples are randomly generated and evaluated using the true rewards in the real environment. Then, AE is applied to transform these high-dimensional policies into lower-dimensional representations. 
Specially, our HNN is a classification model. 
%, so we also need to label each training data as either a promissing policy or a unpromissing policy. 
Therefore, we do not need to predict the fitness of policies like in fitness-based methods. Hence, this approach might be more 
suitable for our framework.
% natural in evolutionary algorithms.
We consider a training data as a promissing sample if its fitness value is greater than the average, and we label it as 1. Conversely, if the fitness value is below the average, we consider it as an unpromissing sample and label it as -1, according to the following equations:\\
\begin{eqnarray}
    \hat{f(y)}=\sum f(y_{i,j})/n,j=1,...,n;
\end{eqnarray}
\begin{equation}
    label(x)=
    \left\{
    \begin{aligned}
        1,f(x)\ge \hat{f(y)}\\
        -1,f(x)<\hat{f(y)}        
    \end{aligned}
    \right.    
\end{equation}
Here, "label" in Equation (4) defines the binary class for each sample based on its fitness. In contrast, "Labels" in Algorithm 1 refers to the collection of these binary labels generated from a batch of training samples. This set serves as the supervision signal for HNN training. 
After completing data preparation and processing, we will split the data-set into training, validation, and testing sets in a ratio of 6:2:2\cite{goodfellow2016deep}. 

% \paragraph{Training the agent model using HNN}

HNN consists of an input layer, hidden layer(s), and an output layer. Next, let’s delve into the specific implementation of HNN.
The input layer, also referred to as the hyperbolic embedding layer, aims to map the input data from Euclidean space to hyperbolic space. This mapping is beneficial for handling nonlinear structured data and can enhance the neural network's ability to learn from manifold data. %In a hyperbolic neural network, 
By mapping the input data to hyperbolic space,
we can better leverage nonlinear operations and standard optimization techniques in hyperbolic space, thereby improving the modeling %and learning 
capabilities for manifold data. The mapping results are processed through a general neural network. 
Correspondingly, the output layer is designed to remap the output data from hyperbolic space back to Euclidean space. 
The overall model formula of the network is as follows:
\begin{eqnarray}
F \oplus_c x=exp^c_0(F(log^c_0(x)))
\end{eqnarray}
%Here we need to explain some basic operations for the conversion between hyperbolic space and Euclidean space:
% \begin{eqnarray}
% \label{for6}
%       x \oplus_c y=& \frac{(1+2c\langle x, y \rangle + c\|y\|^{2})x+(1-c\| x\|^{2})y}{1+2c\langle x,y\rangle}\nonumber\\
%      ~&+c^2\|x\|^{2}\|y\|^2
% \end{eqnarray}

\begin{eqnarray}
\label{for6}
      x \oplus_c y=& \frac{(1+2c\langle x, y \rangle + c\|y\|^{2})x+(1-c\| x\|^{2})y}{1+2c\langle x,y\rangle}+c^2\|x\|^{2}\|y\|^2
\end{eqnarray}
\begin{eqnarray}
\label{for7}
    exp^c_{x}(\nu) = x \oplus_c (tanh(\sqrt{c}\frac{\lambda^c_{x}\|\nu\|}{2})\frac{\nu}{\sqrt{c}\|\nu\|})
\end{eqnarray}
\begin{eqnarray}
\label{for8}
    log^c_{x}(y) =\frac{2}{\sqrt{c}\lambda^c_{x}}arctanh(\sqrt{c}\|-x \oplus_c y\|)\frac{-x\oplus_c y}{\|-x \oplus_c y\|} 
\end{eqnarray}
Eq.\ref{for6} defines the Möbius addition of x and y, Eq.\ref{for7} defines the exponential mapping $Exp_x$ for projecting a vector $\nu$ from the tangent space $T_xM$ at point $x$ to a point $Exp_{x}(\nu)$ $\in$ $M$  on the manifold $M$, and Eq.\ref{for8} is the inverse operation of the exponential mapping, used to project a point on the manifold to the tangent space at another point. 
The parameter c 
%in these equations 
% represents an additional parameter compared to the original Euclidean space, and it 
characterizes the degree of proximity to the center position.
% in the distance. 
In  original Poincaré disk, the radius is equal to $\frac{1}{\sqrt{c}}$. 
When ${c=0}$, these equations degenerate to the Euclidean space. 

Here, we use the Poincaré ball model rather than the Poincaré disk model. 
The Poincaré disk model is usually applied to two-dimensional hyperbolic geometry problems and is more suitable for visualization. 
In high-dimensional policy space, many datasets inherently exhibit multiple coexisting hierarchical structures, 
which cannot be effectively represented in two-dimensional embeddings. 
The Poincaré ball model, on the other hand,  extends to three or higher dimensions,
making it more suitable for handling high-dimensional data, 
% and hyperbolic embedding scenarios, 
such as graph data or neural network embeddings\cite{sala2018representation,ganea2018hyperbolic}.
% High-dimensional embeddings help alleviate optimization challenges. 
In our algorithm, 
the Poincaré ball space is utilized for classifying the superiority or inferiority of policies.
The distribution of policy points in the hyperbolic space reflects differences in policy performance. 
In the Poincaré ball model, 
the additional dimensions allow for more accurate preservation of distances between policy points, enhancing the distinction between ``good" and ``bad" policies\cite{cetin2022hdrn}.

% HNN uses ReLU as the activation function for the hidden layer. ReLU is a type of non-linear activation function, making it more suitable for HNN. Additionally, due to the sparsity of gradients in ReLU, it can enhance the network's generalization ability and reduce overfitting issues. Lastly, ReLU's computational speed is faster, making it easier to implement. 

The output layer of HNN serves the opposite purpose of the input layer, mapping the output from the manifold space back to the original space. 
By using the softmax function as the activation function, the output of HNN is transformed into a probability distribution that clearly indicates the classification probabilities for each category.
This is particularly suitable for classification models and enhances their interpretability.
Because the Poincaré ball possesses a Riemannian manifold structure, we utilize the Riemannian stochastic gradient descent (RSGD) optimizer provided in\cite{bonnabel2013sgd}.

% We train the HNN using the labeled parent data, which has been dimensionally reduced using the AE.
% Then, we input the three randomly perturbed and dimensionally reduced candidate samples into the HNN for pre-selection.

%\paragraph{Using the agent model for pre-selecting RL policies}
Although the HNN is trained using binary labels (i.e., promising or unpromising), the model outputs a continuous probability score indicating the likelihood of each policy belonging to the promising class. During testing, these predicted probabilities are used to establish a soft ranking over candidate policies. This allows the selection process to retain ranking information implicitly, without requiring explicitly ordered labels during training. Thus, the surrogate functions not only as a binary classifier but also as a probabilistic ranker, enabling effective policy filtering based on score-based pre-selection.

A point worth noting is that due to the scarcity of policy samples, we continue to train HNN based on the model trained in the previous 
generation each time. 
Specifically, we initialize an HNN model and save its parameters at the beginning. 
3*N
At each round of evolutionary reinforcement learning,
N parent samples are generated
and returned to RL side for true function evaluation.
Instead of training a new model from scratch every time,
the HNN  model is further updated using  
these evaluated samples.

\subsection{Algorithm Workflow}
The pseudo-code for the concrete algorithm is given in Algorithm \ref{AE-HNN-NCS}. First, perform the initialization operation (lines 1-5), including initializing $N$ randomly generated initial policies, an autoencoder $m_a$, a hyperbolic neural network $m_h$, generating $N$ sub-processes using OpenMPI, and conducting initial evaluations of these $N$ initial policies using the evaluation function $f$ of RL. Then, it enters the external while loop (line 6), which includes sampling $M$ candidate offspring from a random Gaussian distribution (line 8), using $m_a$ as the PE module to map candidate offspring from high-dimensional encoding to low-dimensional (line 9), and using $m_h$ as the surrogate-assisted module to pre-select the candidate offspring with the highest probability of good classification as the true offspring (line 10), and mapping the selected true offspring back from low-dimensional to original high-dimensional (line 11). Subsequently, the evaluation scores $f$ and diversity $d$ of the offspring and parent are calculated, and if match $f(x'_i)+\varphi \centerdot d(x'_i)>f(x_i)+\varphi \centerdot d(x_i)$, it is updated; otherwise, it remains unchanged (lines 12-16). This is the key concept of NCS that trade-off the exploitation and exploration with $f$ and $d$. Finally, the 1/5 success rule is used for updates $\sum_i$. The termination condition of the algorithm is that the current number of executed frames exceeds the total number of frames.
 \begin{algorithm} 
    %\footnotesize 
    \LinesNumbered
    %\SetAlgoNoLine
    \caption{Procedure of AE-HNN-NCS} \label{AE-HNN-NCS}   
    %\SetKwData{localTimerOfGroups}{localTimerOfGroups}
    %\SetKwData{globalTimer}{globalTimer}
    \SetKwData{Q}{Q}
    %\SetKwData{y}{y}
    \SetKwInOut{Input}{input}\SetKwInOut{Output}{output}
    \Input{Number of sub-process $N$, \\
 RL simulator $f$, \\
 Embedding dimension $m$, \\
 Origin policy dimension $n$, \\
 Evaluation limitation $max\_steps$, \\
 Number of candidate policies $M$
 }
    \Output{Final best policy $x^*$}
    \BlankLine  
    Initialize $N$ policies $x_i \in R^n$ uniformly, $i = 1,..., N$;\\
    Initialize an Autoencoder $a$: $n \to m$;\\
    Initialize a Hyperbolic neural network $h$;\\
    Initialize OpenMPI with $N$ sub-processes;\\
    Evaluate the $N$ policies with RL simulator $f$;\\
%        $t \leftarrow 0$; \\
        \Repeat{$steps\_passed \geq max\_steps$}{
        \For{$i$ to $N$}{
            Sample candidate policies {$x'_{i,1},x'_{i,2}...,x'_{i,M}$} from the distribution $p_i \sim N(x_i,\sum_i)$;\\
            Use $a$ to encode candidate policies {$x'_{i,j}$} to {$y'_{i,j}$} where $x'_{i,j} \in R^n$ and $y'_{i,j} \in R^m$;\\
            Use $h$ to choose a best policy {$y^*_i$} with the highest probability of belonging to the good category, as shown in Algorithm 2;\\
            Map {$y^*_i$} to {$x^*_i$} and evaluate {$x^*_i$} with RL simulator $f$;\\
            Calculate $d(p_i)$ and $d(p'_i)$ where $p'_i \sim N(x'_i,\sum_i)$;  //d denotes the diversity \\
            \If{$f(x'_i)+\varphi \centerdot d(x'_i)>f(x_i)+\varphi \centerdot d(x_i)$ \\//$\varphi$ denotes the trade-off parameter}
            {Update $x_i=x'_i,p_i=p'_i$;}
            Update $\sum_i$ according to the 1/5 successful rule;\\
            }   
        }
%       Copy top $|U_t|/2$ solutions of $U_t$ to $P_{t+1}$
    \end{algorithm}

\section{Experimental Setup}\label{sectionExperimentalSetup}
\subsection{Environment and Algorithm Setting}
The Atari 2600 game environment, widely used for benchmarking RL algorithms, is employed to evaluate the performance of our proposed method. As a core component of the Arcade Learning Environment (ALE)\cite{bellemare2013ale}, Atari 2600 provides a diverse suite of challenging games characterized by high-dimensional visual inputs and varied task structures. These tasks include obstacle avoidance (e.g., Highway, Enduro), shooting (Beamrider, Space Invaders), maze navigation (Alien, Adventure), ball games (Pong), platformers (Montezuma’s Revenge), and sports (Bowling, Double Dunk), among others.

Atari games serve as a rigorous testbed for reinforcement learning algorithms due to their unique challenges. First, they feature high-dimensional state representations in the form of raw pixel observations, requiring efficient visual processing capabilities. Second, they are partially observable—agents receive limited information about the game state, necessitating decision-making under uncertainty as they progressively uncover more information through interaction. Third, Atari games often exhibit delayed reward structures, where agents must execute extended sequences of actions before receiving meaningful feedback. This characteristic increases the complexity of learning and demands effective long-term planning.

To comprehensively assess our algorithm’s performance, we selected 10 representative games from the Atari suite. These games are implemented within ALE and provided as standardized interfaces compatible with the OpenAI Gym API\cite{brockman2016gym} standard\footnote{https://github.com/openai/atari-py}.

%To ensure clarity and reproducibility, we summarize the key hyperparameters and design configurations used in our experiments:

%\begin{itemize}
%\item \textbf{Policy Networks:} For Atari, we use convolutional neural networks with 2.3M parameters; for Mujoco, we use 3-layer MLPs with ReLU activations and parameter sizes ranging from 10k to 50k.
%\item \textbf{Embedding Dimension:} AE reduces policy representations to a 10-dimensional latent space.
%\item \textbf{Population Size (M):} Each generation uses a population of 3 candidate policies.
%\item \textbf{AE Architecture:} 4-layer encoder/decoder with 256-128-64-10 units, trained with MSE loss and ReLU activations.
%\item \textbf{HNN Structure:} 2-layer hyperbolic MLP with curvature $c = 1$, trained using binary cross-entropy.
%\item \textbf{Training Details:} AE and HNN are updated every generation using a batch of 8M samples, learning rate = 1e-3, batch size = 256.
%\end{itemize}}

\subsection{The Compared Algorithms}
%出现过的算法需要在前文中提到，第一次出现用全称
To demonstrate the effectiveness of our algorithm, we compare it with several classical and state-of-the-art RL methods. Specifically, our evaluation includes fundamental RL baselines such as A3C\cite{mnih2016async} and PPO\cite{schulman2017ppo}. Both them are widely adopted gradient-based RL algorithms that optimize neural networks through traditional backpropagation.
%The proposed algorithm is compared with several classical and state-of-the-art RL methods to demonstrate its effectiveness. Our comparisons include basic RL baselines such as A3C\cite{mnih2016asynchronous} and PPO\cite{schulman2017proximal}. A3C and PPO are popular gradient-based reinforcement learning methods that train networks using traditional backpropagation.

Additionally, we also compare our algorithm with ERL-related baselines, including CES\cite{chrabaszcz2018back}, NCS\cite{tang2016ncs}, and PE-SAERL\cite{tang2022pe}. 
CES is a variant of evolutionary policies that optimizes the weights of policy neural networks to enhance learning performance. Its core principle involves updating these weights using the natural gradient method.
NCS, which is integrated into our algorithm framework, introduces negative correlation into evolutionary algorithms to improve exploration of the solution space and enhance the search for optimal policies. 
PE-SAERL is the first algorithm to integrate a dimension reduction module into the ERL framework, which makes it a useful benchmark  for evaluating our approach.

Furthermore, to highlight the advantages of our algorithm, we compare it with two state-of-the-art ERL algorithms: PLASTIC\cite{lee2024plastic} and STORM\cite{zhang2023storm}. 
PLASTIC improves sample efficiency in RL by enhancing input plasticity through loss landscape optimization and fine-tuning gradient propagation, thereby improving label plasticity. Experimental results indicate that PLASTIC is highly competitive, underscoring the importance of maintaining model plasticity to enhance sample efficiency. 
STORM, in contrast, introduces random noise and integrates the sequence modeling capabilities of Transformers with the stochastic nature of variational autoencoders, enabling the development of an efficient world model that mitigates discrepancies between the learned model and the real environment. 

By benchmarking our algorithm against these state-of-the-art approaches, we aim to demonstrate its effectiveness in addressing the challenges of sample efficiency and dimensionality reduction in ERL.
\subsection{Performance Metrics}
The quality of the policy is evaluated based on its test score, which represents the average cumulative reward obtained over 30 or more game repetitions without any framework constraints. The test score is defined as the total rewards accumulated by a policy across multiple game episodes. Typically, a game episode starts with a random ``noop" (i.e., taking no action) frame and and continues until the game signals termination\cite{bellemare2013ale}. 
During each episode, the agent selects actions based on the current game state and interacts with the environment. At each time step, the game environment provides the agent with an immediate reward based on the agent's action and the game state. 
To obtain reliable performance estimates, the test score is computed as the average reward over multiple episodes, typically 30. This approach ensures that the agent's performance is assessed across a diverse range of game scenarios.
Additionally, we impose an upper limit of 100,000 frames per episode. This constraint ensures that the agent has sufficient opportunities to explore various actions and state transitions while preventing it from becoming stuck in a "deadlock" situation.
%The test scores are calculated from multiple game episodes. A common approach is to play a certain number of episodes (e.g., 30) and then compute their average scores. This method allows us to evaluate the performance of the agent over a variety of game scenarios.
%We limit the length of an episode to a very large value of 100,000 frames. By restricting the length of episodes, we ensure that the agent has the opportunity to try out a variety of actions and state transitions within a certain time frame, thus avoiding getting stuck in a "deadlock". 

It is important to note that for ERL methods (i.e., comparative algorithms CES, NCS, and PE-SAERL), the total game frames are set to 25 million, while for gradient-based methods (i.e., baseline PPO, A3C), the total game frames are set to 10 million. This discrepancy arises because gradient-based methods rely on backpropagation, and using the same number of game frames for both approaches would lead to an unfair comparison. 
Salimans et al.\cite{salimans2017evolution} under identical experimental conditions and computational runtime, the ratio of game frames consumed by ERL methods to those used by gradient-based methods is approximately 2.5. This observation justifies the difference in frame allocation. Further details on the experimental settings can be found in Appendix A.
\begin{table*}[htbp]
\tabcolsep0.03in
\renewcommand{\arraystretch}{1.2}
\caption{\textcolor{black}{
Performance results of the Random, Human, baseline and State-of-the-art ERLs on ten Atari games.}}
%The first column means the test instance number (For example, 1 corresponds to LSMOP1)
%and the second column corresponds to  the search space dimension, hereafter.}
\label{tableresult}
\centering
\resizebox{\textwidth}{!}{
  \tabcolsep0.01in
  \renewcommand{\arraystretch}{1.2}
% \scriptsize
% \small
\begin{tabular}{c c| c c| c c c c c| c c c}
%可以多写几个c,超了也不影响表格正确性的
\toprule
Game &Performance  &  Random &  Human &CES &PPO &A3C &NCS &PE-FCPS-NCS    &PLASTIC  & STORM  & AE-HNN-NCS     \\ 
\midrule 
\multicolumn{2}{c|}{Time} &40M&40M&0.1B&40M&40M&0.1B&0.1B&40M&40M&0.1B\\\hline
\multirow{3}*{Alien}    
& Average   &   230.0  &   7190.0  & 562.5& 632.7& 771.7& 989.2& 825.7& 1032.0  & 984.7   & \hl{1893.5}  \\ 
& Best   &   239.0  &   7930.0  &640.5& 800.7& 798.0& 1297.5& 1054.7& 1127.3  &798.0  &  \hl{2577.9} \\ 
& Percent   &   12.1\%  &   -& 29.7\%&33.4\%&40.7\%&52.5\%&42.6\%  & 54.5\%  &40.7\%&  100.0\%  \\ 
\hline
\multirow{3}*{Bowling}    
& Average   &   23.2  &  157.9&   58.5  &   23.2   & 0.0    & 63.2  &  91.3    & 21.0  & 60.5     &  \hl{99.7}  \\ 
& Best   &   23.9  &  165.3&   63.8  &   23.9   &0.0  & 70.2  &  107.5  & 27.9  &69.3   &  \hl{123.2} \\ 
& Percent   &   23.2\%  &   -  &   58.7\%  &   23.3\%    &0.0\%& 63.3\%  &  91.6\%& 21.0\%  &60.6\%&  100.0\%  \\ 
\hline
\multirow{3}*{DoubleDunk}    
& Average   &   -18.9  &   -16.5  &   -3.7  &   -3.1    & -2    & -1.42  &  -0.6  & -2.7  & -9.65    &  \hl{-0.23}  \\ 
& Best   &   -18.6  &   -15.9  &   -2.8  &   -3.0    &-1.5  & -0.8  & 0.0  & -0.39  &-1.5   &  \hl{0.0} \\ 
& Percent   &   0.0\%  &   -  &   0.0\%  &   17.3\%    &48.9\%& 65.7\%  &  89.3\%& 86.7\%  &49.5\%  &  100.0\%  \\ 
\hline
\multirow{3}*{Freeway}    
& Average   &   0.0  &   27.3  &   14.2  &   14.4    & 0.0    & 16.5  &  21.3 & 25.7  & 26.0    &  \hl{26.1}  \\ 
& Best   &   0.0  &   29.7  &   15.9  &   22.0    &0.0  & 17.1  & 22.8  & 27.1  &26.9  &   \hl{27.8} \\ 
& Percent   &   0.0\%  &   -  &   61.5\%  &   62.3\%    &0.0\%& 71.4\%  &  92.2\%  & 98.4\%  &99.6\%  &  100.0\%  \\ 
\hline
\multirow{3}*{BeamRider}
& Average & 265.8 & 16925.7 & 442.5 & 615.4  & 643.2 & 729.7 & 778.1 & 798.6 & 851.3  & \hl{977.1}  \\
& Best & 269.3 & 17302.1 & 509.7 & 649.5  & 882.1& 782.6 & 818.2 & 8263.9 & 882.1&  \hl{1010.4}  \\
& Percent & 26.9\% & - & 45.2\% & 62.9\%  & 65.8\% & 74.6\% & 79.6\% & 81.7\% & 87.1\%  & 100.0\%  \\
\hline
\multirow{3}*{Pong}
& Average & -20.9 & 4.9 & -7.3 & -21.5  & -23.7 & -9.7 & -3.8& -17.7 & 1.1  & \hl{2.3}  \\
& Best & -18.7 & 8.1 & 5.2 & -19.7  & -23.7 & -9.1 & 2.7& -15.3 & 3.7  & \hl{8.7}  \\
& Percent & 0.0\% & - & 63.0\% & 8.4\%  & 0.0\% & 53.8\% & 76.5\% &13.8\% & 94.8\%  & 100.0\%  \\
\hline
\multirow{3}*{Enduro}
& Average & 0.0 & 178.5 & 5.8 & \hl{258.7}  & 0.0 & 7.9 & 10.5 & 42.8 & 0.0  & 25.8  \\
& Best & 0.0 & 215 & 8.1 & \hl{476.2}  & 0.0 & 11.2 & 12.0&47.3 & 0.0  & 27.3  \\
& Percent & 0.0\% & - & 2.2\% & 100.0\%  & 0.0\% & 3.1\% & 4.1\% & 16.5\% & 0.0\%  & 10.0\%  \\
\hline
\multirow{3}*{SpaceInvaders}
& Average & 148 & 1670.2 & 387.2 & 453.7  & 595.9 & 953.2 & 972.0& 484.5 & 809.3 & \hl{1139.5}  \\
& Best & 152.0 & 1678.0 & 540.0 & 525.8  & 827.1 & 1047.0 & 1072.3 & 523.6 & 827.1 &  \hl{1203.0}  \\
& Percent & 12.9\% & - & 33.9\% & 46.1\%  & 52.2\% & 83.6\% & 85.3\% & 42.5\% & 71.1\%  & 100.0\%  \\
\hline
\multirow{3}*{Venture}
& Average & 0.0 & 937.5 & 320.0 & 0.0  & 0.0 & 498.0 & 562.0& 430 & 0.0  & \hl{696.0}  \\
& Best & 0.0 & 1025 & 415.0 & 0.0 & 0.0  & 577.0 & 643.0 & 459.3 & 0.0 &  \hl{739.0}  \\
& Percent & 0.0\% & - & 45.9\% & 0.0\%  & 0.0\% & 71.6\% & 80.7\%& 61.8\% & 0.0\%  & 100.0\%  \\
\hline
\multirow{3}*{MontezumaRevenge}
& Average & 0.0 & 1250.6 & 0.0 & 10.6  & 0.0 & 2.7 & 3.9 & 11.2 & 8.1  & \hl{28.7}  \\
& Best & 0.0 & 1287.3 & 0.0 & 11.8  & 0.0 & 5.3 & 7.8 & 13.7 & 9.3  & \hl{63.5}  \\
& Percent & 0.0\% & - & 0.0\% & 36.9\%  & 0.0\% & 9.4\% & 13.5\% & 39.1\% & 28.2\%  & 100.0\%  \\
%Pong Atlantis Enduro SpaceInvaders Venture MontezumaRevenge

\bottomrule
\end{tabular}}
% \end{scriptsize}
% \begin{tablenotes}
% \end{tablenotes}
%\end{sidewaystable}
\end{table*}
\section{Results and discussions}\label{sectionExperimentalResults}
%%%%%%%%%%%%%%%%%%%%%%%%%%%%%%%%%%%%%%%%%%%%%%%%%%%%%%%%%%%%%%%%%%
\subsection{General Performance}
For each Atari game, we trained a neural network policy model and performed 20 repeated tests\cite{tang2022pe}
on the trained model to obtain the average test score as the performance metric. The test scores for each algorithm on these games are shown in Table \ref{tableresult}. 
The average row displays the average performance over the ten runs, with the highest score in bold. 
The percentage row represents the average performance as a percentage of the best performance. 
From the table, it is evident that our algorithm generally outperforms all the compared methods.
In the Bowling and Freeway games, RL baselines performed relatively poorly, while ERL showed better performance, with our algorithm performing the best. Notably, in the Bowling game, our algorithm showed significant improvement in the highest score achieved.
In the DoubleDunk game, all algorithms obtained negative scores, indicating failure of the agent. However, relatively speaking, our algorithm performed the best as its average score is closest to 0. 
In the Enduro game, although our algorithm didn't achieve the highest performance, it was only surpassed by PPO. 
In some specific tasks, the application of the agent might lead to exploration in worse directions, but further exploration is needed to identify the specific reasons. Overall, our algorithm generally outperformed the other algorithms in terms of average test scores, indicating its effectiveness in Atari game playing.

\begin{figure}[ht]   
   \centering
     \includegraphics[width=250pt]{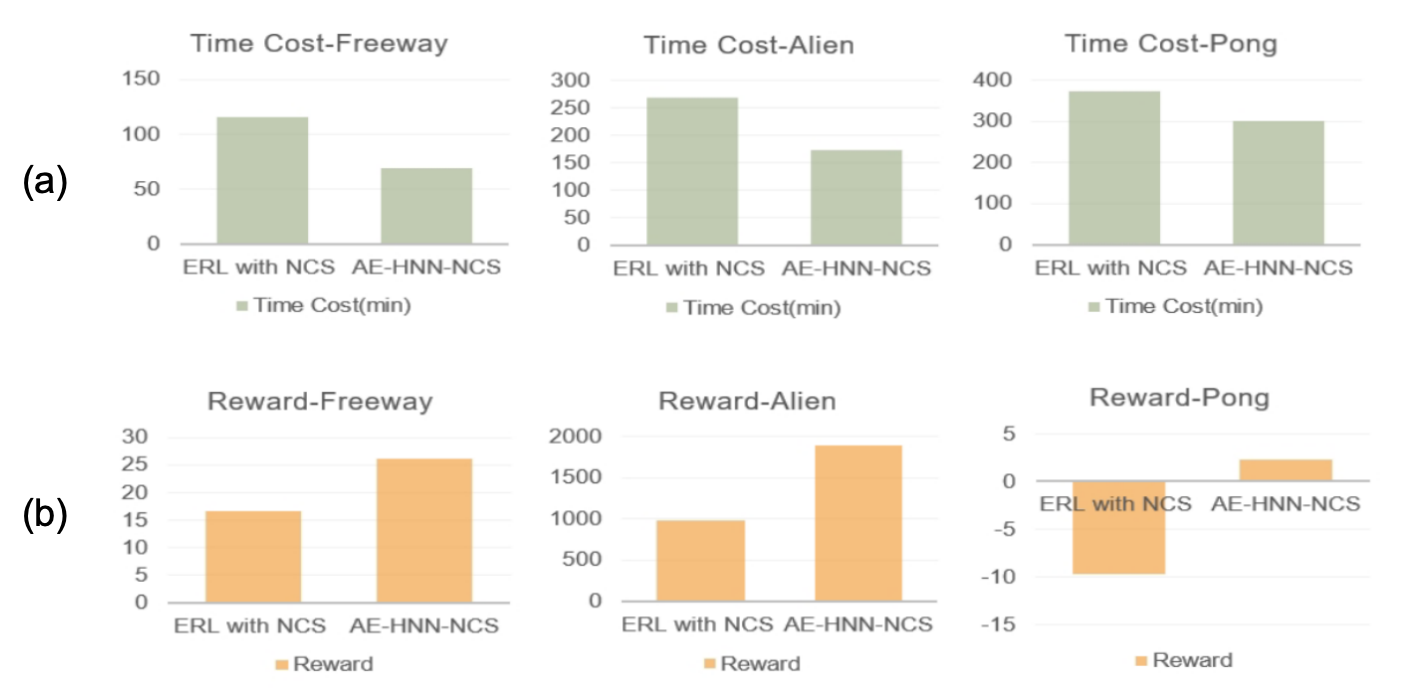}  %%%lbd20150622 %%%lbd20160114
   \caption{Performance Analysis of ERL with NCS and AE-HNN-NCS in Different Environments.
   Figures (a)and (b) show the comparison between ERL with NCS and AE-HNN-NCS in terms of time cost and final policy reward.
   % \textcolor{black}{Framework or Algorithm?}
   }
   \label{case}
\end{figure}

As shown in Fig.\ref{case}, our algorithm demonstrates a clear advantage over traditional ERL, not only in terms of final reward but also in addressing the critical issue of low search efficiency. %Specifically, the graphs in examples (a) highlight that our algorithm significantly reduces time overhead compared to traditional ERL. 
To assess the trade-off between the added cost of training AE and HNN and the savings from reduced real evaluations, we conducted a detailed runtime analysis shown in Fig. 3(a). While AE and HNN introduce approximately 15\% extra computation per generation, the number of real environment evaluations is reduced to nearly one-third. As a result, the overall wall-clock training time is reduced by 38\% on average. This confirms that the computational overhead is more than compensated by the gains in sample efficiency.
This advantage primarily comes from the pre-selection mechanism introduced by the HNN. By classifying the offspring generated by the parent, HNN pre-selects samples with the highest likelihood from the good classifications, thereby reducing the need for real sample evaluations. Since only the final selected single sample requires actual evaluation, time efficiency is significantly improved.

\subsection{Ablation Study}
To demonstrate the effectiveness of two newly added modules, we conducted ablation experiments on AE and HNN. We separately added the AE and HNN to the original PE-SAERL, and ran 20 repeated tests on different games to obtain the average test scores as performance metrics. The results of the ablation experiments on these games are presented in Table \ref{tableAblation}.

%First of all, we can observe that the experimental results with the addition of the autoencoder are better than those of the original framework. The original framework used Random Embedding (RE) as the dimensionality reduction module, which is a simple method that achieves data reduction through straightforward random projection without involving complex feature learning or data compression. Moreover, it cannot capture the intrinsic structure or meaning of the data, thus may not provide the optimal data reduction representation, especially when critical features or semantic information exist in the data. On the other hand, autoencoders not only enable data reduction but also capture the meaningful representation of the input data by learning how to reconstruct it. They help discover important features and structure within the data. Additionally, autoencoders exhibit robustness to noise in the input data and can learn to retain key information, giving them a certain degree of denoising capability.
Based on the table, we can observe that the experimental results after adding an AE are better than those of the original framework. The original framework used Random Embedding (RE) as the dimensionality reduction module, which reduces data dimensions through simple random projection, without involving complex feature learning or data compression. 
Additionally, RE cannot capture the intrinsic structure or semantics of the data, and when critical features or semantic information exist, RE may fail to provide the best dimensionality reduction representation. 
In contrast, autoencoders not only reduce the data dimensions but also capture meaningful representations by learning how to reconstruct the input data, enabling the discovery of key features and structures in the data. Furthermore, autoencoders show a certain degree of robustness to noise in the input data, allowing them to learn and retain key information, giving them some denoising capability.
As shown in Fig. \ref{tsne}, the analysis of the t-SNE plot reveals that after dimensionality reduction using AE, the final optimal policies exhibit a trajectory-like clustering in the low-dimensional space.
%Through the analysis of the t-SNE plot, shown as Fig.\ref{tsne} we found that after using AE for dimensionality reduction of high-dimensional policies, the final optimal policies showed a trajectory-like clustering in the low-dimensional space. 
This suggests that the search trajectory of the policy is more orderly, and the data distribution is more compact in the low-dimensional space. The autoencoder not only reduces the dimensionality but also retains important features of the policies, helping them explore more orderly and converge toward the optimal solution in the low-dimensional space. 
In contrast, the policies obtained with PE were scattered across the exploration space without a clear clustering trend, indicating that PE failed to effectively capture the correlation and structure between policies, leading to lower exploration efficiency.

Similarly, we observe that using HNN as the surrogate-assisted module performs better than traditional surrogate models. The main reasons can be summarized as follows: First, HNN has learning capabilities and can be trained using algorithms like backpropagation to adapt to different classification tasks. In contrast, FCPS typically relies on fixed nearest neighbors for classification and lacks learning capabilities. 
Second, HNN can learn complex nonlinear relationships in the data and handle non-Euclidean data, making it suitable for classification tasks in complex feature spaces. On the other hand, FCPS and NN rely on Euclidean geometry, making it difficult to effectively capture the inherent structure and nonlinear relationships of complex data. Additionally, FCPS and NN tend to perform better in simple or linearly separable feature spaces, while they may exhibit weaker performance in high-dimensional or complex data spaces. 
Finally, HNN demonstrates better generalization ability in high-dimensional or sparse data scenarios, requiring fewer samples, whereas FCPS and NN may suffer from the curse of dimensionality, limiting their ability to represent data in high-dimensional spaces.

Additionally, we have also found that replacing the PE module with AE performs better than replacing the surrogate-assisted module with HNN. This may be because AE can retain the important features of the data, which has a greater impact on the effectiveness of the classification algorithm. As a result, even FCPS can demonstrate relatively good classification results. On the other hand, RE is unable to capture the important features of the data, so even with the use of HNN, accurate learning cannot be achieved, resulting in less satisfactory outcomes. Of course, the combination of AE and HNN demonstrates the best performance, as each module contributes its unique strengths.

\begin{table*}[htbp]
\tabcolsep0.03in
\renewcommand{\arraystretch}{1.2}
\caption{\textcolor{black}{
Ablation results.}
%\textcolor{black}{\\similar with parameter analysis,
%speedup.
%This table 
%is critical
%since it 
%supports
%your major
%contributions
%\\
%Besides,
%in your next work,
%experimental 
%runtime
%should be taken into 
%consideration.
%\\
%i.e. 
%your next work title better be xxx
% method for High-efficiency ERL.
%}
%\textcolor{black}{\\
%20240228\\
%Ablation and parameter experiments only need part of all the test cases,
%instead of all the test cases.\\
%I suppose half or 2/3 will be ok,
%as soon as the conclusions is not out of our expectation.\\
%You can choose a fastest way to finish the experiments in this table.\\
%}
}
%The first column means the test instance number (For example, 1 corresponds to LSMOP1)
%and the second column corresponds to  the search space dimension, hereafter.}
\label{tableAblation}
\centering
\resizebox{\textwidth}{!}{
  \tabcolsep0.01in
  \renewcommand{\arraystretch}{1.2}
% \scriptsize
% \small
\begin{tabular}{c c |cc |cc    |c c c}
\toprule
Game &Perforcemence  &  PE-FCPS-NCS &  AE-FCPS-NCS &PE-NN-NCS  & PE-HNN-NCS  &HNN- NCS &AE-HNN-NCS(disk) &AE-HNN-NCS(ball)     \\ 
\midrule 
\multicolumn{2}{c|}{Time} &0.1B&0.1B&0.1B&0.1B&0.1B&0.1B&0.1B\\\hline
\multirow{3}*{Alien}    
& Average & 825.7 & 1527.9 & 1093.1 & 1258.3 &1147.3&1672.8& 1893.5  \\
& Best & 1054.7 & 1923.6 &1125.8 & 1671.6 &1537.2&2193.7& 2577.9 \\
& Percent & 43.6\% & 80.7\% & 57.7\%&66.4\% &60.1\%&88.3\%& 100.0\%  \\
\hline
\multirow{3}*{Bowling}    
& Average & 91.3 & 95.2 &92.9 & 93.7 &91.8&97.3& 99.7  \\
& Best & 107.5 & 114.3 & 109.1 & 109.8 &112.9&119.2& 123.2 \\
& Percent & 91.5\% & 95.4\% & 93.1\% & 93.9\% &92.1\%&97.6\%&100.0\%   \\
\hline
\multirow{3}*{BeamRider}
& Average & 778.1 & 877.3 &803.5& 819.6 &874.3&905.6& 977.1  \\
& Best & 818.2 & 917.0 &855.1& 871.4 &887.2&973.1& 1010.4 \\
& Percent & 79.6\% & 89.7\% & 82.2\% & 83.9\% &89.4\%&92.7\%&100.0\%   \\
\hline
\multirow{3}*{Pong}
& Average & -3.8 & -1.4 & -2.9 &-2.3 &1.7&1.9& 2.3  \\
& Best & 1.7 & 4.2 & 3.5& 3.9 &2.4&5.4& 8.7 \\
& Percent & 0.0\% & 39.3\% & 14.7\% & 24.5\%  &73.9\%&82.6\%&100.0\%   \\
\hline
\multirow{3}*{Enduro}
& Average & 10.5 & 22.7 &14.9& 17.1 &15.9&19.4& 25.8  \\
& Best & 12.0 & 25.8 &17.5& 19.1 &18.7&21.8& 27.3 \\
& Percent & 40.1\% & 87.9\% & 57.7\% & 66.2\% &61.6\%&75.2\%&100.0\% \\
\hline
\multirow{3}*{SpaceInvaders}
& Average & 972.0 & 1075.4 & 1001.8 & 1023.7&979.3&1074.7& 1139.5  \\
& Best & 1072.3 & 1128.1 & 1039.1& 1095.9 &1074.2&1157.3& 1203.0 \\
& Percent & 85.3\% & 94.4\% & 87.9\% & 89.8\% &85.9\%&94.3\%&100.0\%   \\
\hline
\multirow{3}*{Venture}
& Average & 562.0 & 623.8 & 571.7& 583.2 &577.4&642.5& 696.0  \\
& Best & 643.0 & 699.0 & 649.8& 654.7 &637.9&715.8& 739.0 \\
& Percent &80.7\% & 89.6\% & 82.1\% & 83.8\% &82.9\%&92.3\%&100.0\%   \\
\hline
\multirow{3}*{MontezumaRevenge}
& Average & 3.9 & 20.3 & 15.8 &19.4&17.9&24.6& 28.7  \\
& Best & 7.8 & 47.6 & 33.2&21.3 &25.4&39.8& 63.5 \\
& Percent & 13.5\% & 70.7\% & 55.1\% & 67.5\% &61.9\%&85.7\%&100.0\%   \\
%Pong Atlantis Enduro SpaceInvaders Venture MontezumaRevenge

\bottomrule
\end{tabular}}
% \end{scriptsize}
% \begin{tablenotes}
% \end{tablenotes}
%\end{sidewaystable}
\end{table*}

Overall, the ablation experiments confirmed the positive impact of both the AE and the HNN on the performance of the original PE-SAERL algorithm across a variety of games.

In fact, we also conducted parameter sensitivity experiments. Due to space limitations, the experiments on the impact of different numbers of candidate solutions on the algorithm's performance are provided in AppendixB.

\subsection{Why Pre-select, not Reproduce, in the Embedded Space?}\label{whypre-select}
\textcolor{black}{To verify the consistency between the rankings of high-dimensional and low-dimensional policies, we conducted experiments on six Atari games (Alien, Freeway, Bowling, Pong, DoubleDunk, Enduro) using three different strategies. Specifically, 
1) strategy 1: Pre-select in the embedded space, while reproduce in the original space; 
2) strategy 2: Pre-select in the embedded space, while reproduce in the embedded space; 
3) strategy 3: Pre-select in the original space, while reproduce in the embedded space. 
To assess which strategy utilizes the embedded space and facilitates the search better, we employed Spearman's Rank Correlation and Kendall's Tau to measure the correlation between the rankings of high-dimensional policies (in terms of fitness) and low-dimensional policies (in terms of HNN). Experimental results show that strategy 1 achieved significantly better consistency than strategies 2 and 3, with strategy 2 outperforming strategy 3.}

\textcolor{black}{Further analysis shows that the superior performance of strategy 1 is due to ranking being performed solely in the low-dimensional space, without involving reproduction. The pre-selected embedded policies are directly mapped back to the original space using hashing mapping, rather than decoding with AE. This results in a perfect, error-free mapping from low-dimensional to high-dimensional. The AE effectively extracts key dimensional information, which directly influences the factors affecting the reward. As a result, the ranking in the low-dimensional space maintains the ordinal relationship of the high-dimensional policies.
In contrast, strategy 2 involves decoding and re-encoding with AE, which introduces reconstruction errors in the high-dimensional space, negatively impacting ranking consistency. However, since the low-dimensional space still retains the local ordinal relationships of the policies, strategy 2 demonstrates better consistency than strategy 3.
Strategy 3, on the other hand, exhibits the worst consistency. The reproduction operators in the low-dimensional space do not preserve the complete information of the high-dimensional policies, leading to decoded high-dimensional policies that fail to accurately reflect their original performance. This disrupts the ranking consistency between the high- and low-dimensional policies.}

\textcolor{black}{From a theoretical perspective, the goal of AE is to capture the primary features of the data by minimizing global reconstruction error while preserving key information. As a result, the low-dimensional representation is well-suited for tasks based on ordinal relationships, such as pre-selection. However, reproduction operators require a complete representation of the individuals, which imposes greater demands on the low-dimensional representation. When the reproduction process fails to fully capture the high-dimensional policies, the performance of the low-dimensional policies in the high-dimensional space may differ from their evaluations in the low-dimensional space, thus impacting ranking consistency. This explains why strategy 1, which ranks policies solely in the low-dimensional space, achieves the best consistency; while strategies 2 and 3, which involve decoding and searching, perform worse. In conclusion, the key information preserved in the low-dimensional space is adequate for ranking tasks but insufficient to support accurate reproduction operations.}

\begin{figure}[ht]   
   \centering
     \includegraphics[width=250pt]{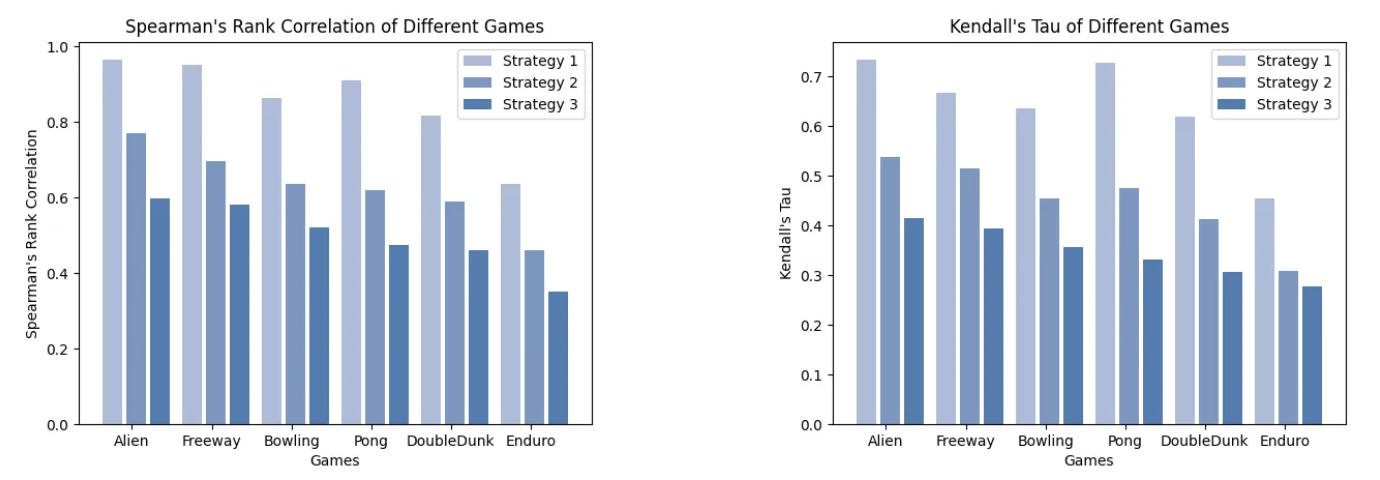}  %%%lbd20150622 %%%lbd20160114
   \caption{The ranking consistency of final performances among three strategies—(1) high-dimensional space searching with low-dimensional space pre-selection, (2) low-dimensional space searching with low-dimensional space pre-selection, and (3) low-dimensional space searching and high-dimensional pre-selection.
   % \textcolor{black}{The text font size
   % in this figure 
   % is too small.
   % you need to edit the original figure file instead of the width in latex
   % to make the font size right.\\
   % }
% \textcolor{black}{font size of lower texts still not good ;
% Information density too low;
%  ask yyt how to draw figures with appropriate font size.}
   }
   \label{cor1}
\end{figure}

\subsection{Visualization of Policy Structure and HNN Pre-selection}
\begin{figure*}[ht]
\centering
\includegraphics[width=400pt]{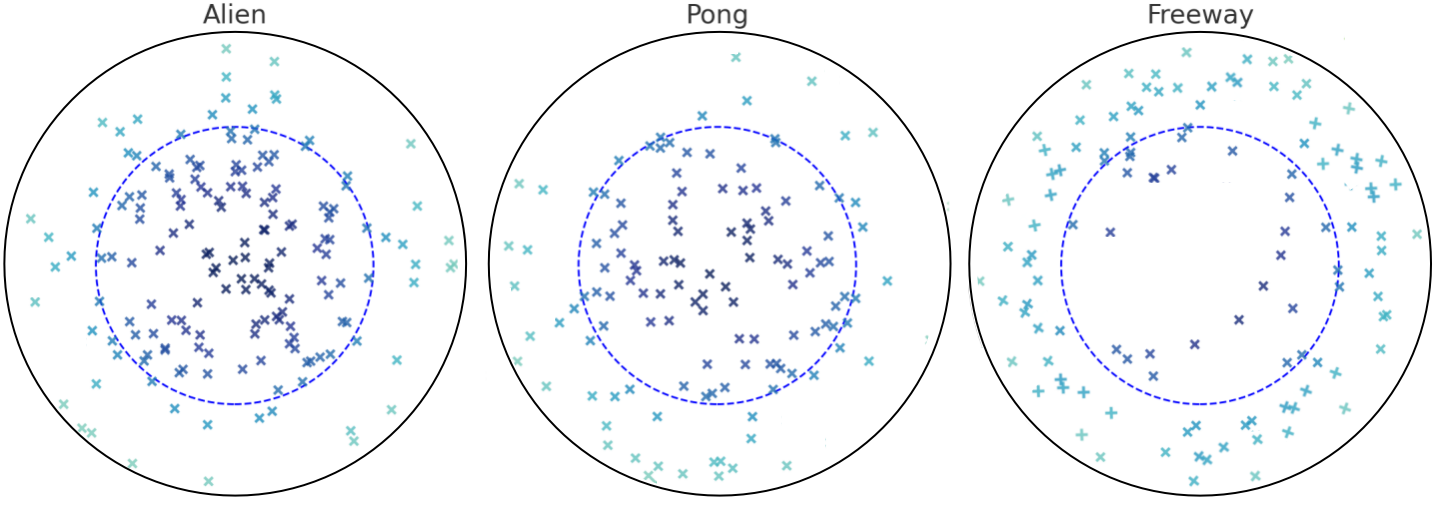}
\caption{The t-SNE visualization results of HNN-based policy pre-selection on three Atari games: Alien, Pong, and Freeway. Each point represents a policy embedding in hyperbolic space; color indicates HNN-predicted score (dark blue = promising); dashed ring marks the high-confidence region used for evaluation.}
\label{fig:hnn_preselection_atari}
\end{figure*}
The fig.\ref{fig:hnn_preselection_atari} visualizes the hyperbolic policy embeddings generated via an autoencoder, where each point represents a candidate policy. The color indicates the HNN-predicted probability of being a high-quality policy. The blue dashed circle highlights the high-confidence region (\(p > 0.5\)) selected for real evaluations.

From the figure, we observe that compared to Pong and Freeway, our method exhibits a more pronounced hierarchical structure and confidence concentration in the Alien environment. This is largely attributed to the inherently hierarchical nature of Alien, which involves multi-stage planning, tactical shooting, and reactive avoidance~\cite{bellemare2013ale, mnih2015dqn}. These results indicate that our hyperbolic modeling approach is particularly suited for environments with structured policy hierarchies.

This visual pattern also aligns with the quantitative results shown in Table \ref{tableresult}: specifically, our method achieves a performance improvement 57.4\% (from 42.6\% to 100\%) on Alien, which is significantly higher than the gain observed on Freeway's 7.8\% (92.2\% to 100\%). This further validates the effectiveness of HNN-based pre-selection in hierarchical tasks.

\subsection{t-SNE Visualization Result Analysis}
%t-SNE是主体

 \begin{figure*}[ht]   
   \centering
     \includegraphics[width=400pt]{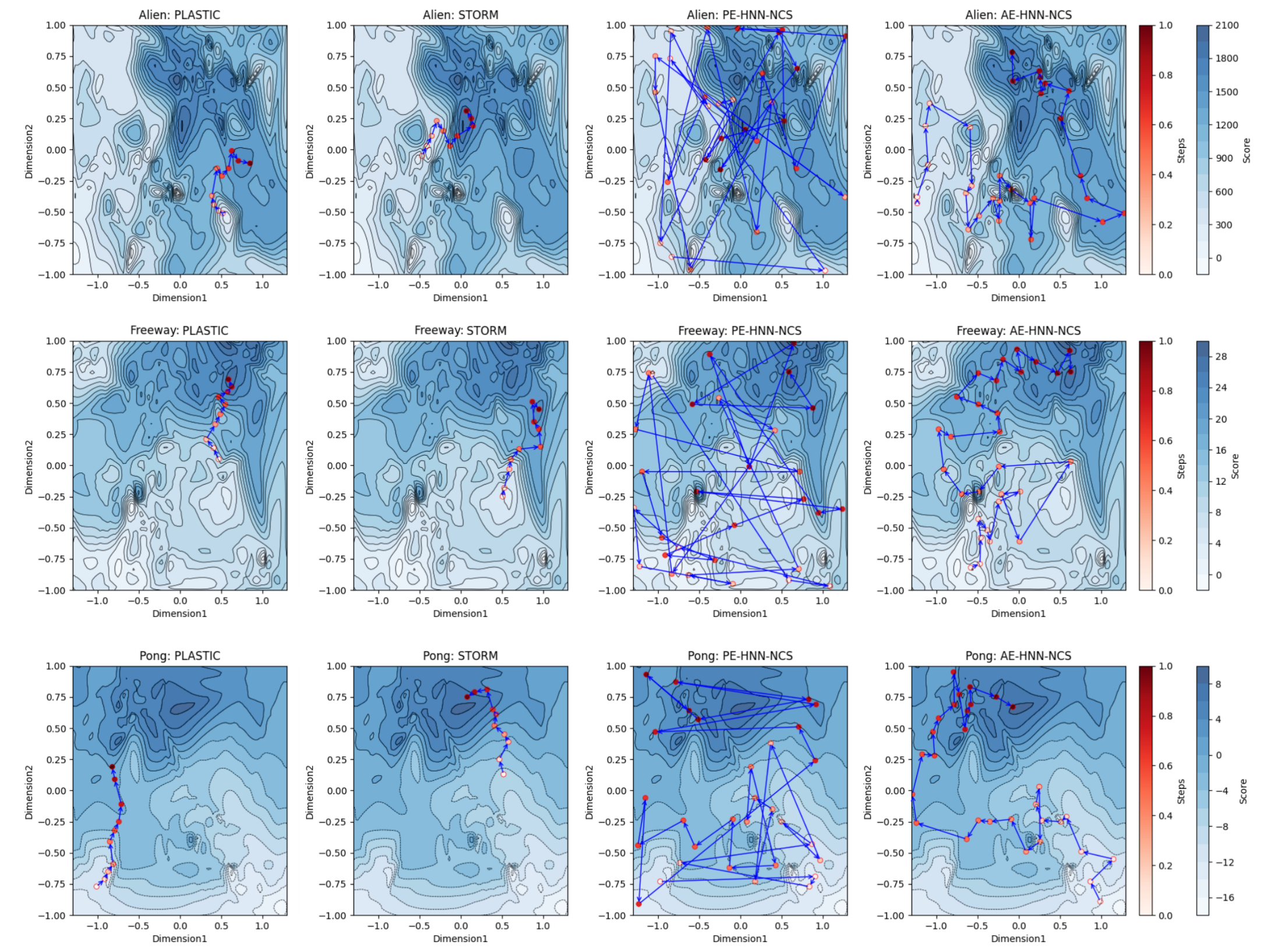}  %%%lbd20150622 %%%lbd20160114
   \caption{The t-SNE visualization results of four algorithms (PLASTIC, STORM, AE-HNN-NCS, PE-HNN-NCS) on three Atari games 
   (Alien, Freeway, Pong). The red node represents the final optimal policy point, and the trajectory line illustrates the search path of the policy during the training process..
   % \textcolor{black}{The text font size
   % in this figure 
   % is too small.
   % you need to edit the original figure file instead of the width in latex
   % to make the font size right.\\
   % }
% \textcolor{black}{font size of lower texts still not good ;
% Information density too low;
%  ask yyt how to draw figures with appropriate font size.}
   }
   \label{tsne}
\end{figure*}
%\begin{figure*}[htbp]   
%   \centering
%     \includegraphics[width=450pt]{pdf/Atari.png}  %%%lbd20150622 %%%lbd20160114
%   \caption{\textcolor{black}{The visualization implementation of Atari game Bowling on three Algorithm {AE-HNN-NCS,STORM,PLASTIC}}.
   %\textcolor{black}{These figures 
   %are beautiful and fancy.\\ Yet What
   %information can we obtain 
   %from these figures?\\
   %Does it show the advantage 
   %of our algorithm?\\}
%   \textcolor{black}{If possible,
%   Try to cite the following
%papers in the main body of this paper (Use  sentences
%different from the following):\\
%\cite{li2023tevc}\\
%In the future,
%Forth, we need to test the performance of RM-SAEA on  complicated
%real-world and/or combinatorial problems 
% such as cloud computing and the traveling salesman problem
%\cite{yangp2024reduce,hong2020efficient,hong2019scalable,liu2023how}.\\
%Finally, we would like to 
%theoretically analyze the proposed method %\cite{qian2011analysisaij2013,bian2023stochasticijcai}.\\
%}
%} 
%\label{atari}
%\end{figure*} 
 To demonstrate the contribution of the AE module in our algorithm to performance, we have conducted visualization experiments on four algorithms in three games. The t-SNE plots are as  fig.\ref{tsne}. 
 The STORM algorithm leverages the powerful sequence modeling capability of Transformers combined with random noise to enhance the adaptability of policies. 
 The PLASTIC algorithm, on the other hand, focuses on improving the sample efficiency of reinforcement learning by enhancing the plasticity of the model. It achieves this by integrating multiple techniques that improve the plasticity of both inputs and labels. 
 Our algorithm AE-HNN-NCS primarily emphasizes dimensionality reduction through the AE module, along with the advanced learning and representational power of the HNN, particularly in handling complex nonlinear data, to boost policy performance in challenging tasks. Finally, PE-HNN-NCS is the baseline algorithm.
 
 We selected three games with distinct characteristics—Alien, Freeway, and Pong—as experimental samples. After analyzing the performance of AE-HNN-NCS, STORM, and PLASTIC in these three games, it can be found that AE-HNN-NCS shows the best performance.
 %In the Alien game, policy effectiveness depends on the player's ability to quickly adapt to different levels, understand enemy movement patterns, and react swiftly under adverse conditions. The game also features a sparse reward mechanism, which poses significant challenges for reinforcement learning (RL) algorithms\cite{bellemare2016unifying}. Given these conditions, PLASTIC's input plasticity and label plasticity allow it to better adapt to the ever-changing game environment and dynamic input-output relationships. As a result, it generates more consistent and effective policies over multiple game rounds, contributing to improved sample efficiency. On the other hand, STORM, which relies on imagined environment data for policy training, may lead to policy inconsistency and accumulate imagination errors. Therefore, we observed that policies generated by PLASTIC were more stable, clustered, and performed better overall.
 Alien-like problems have complex characteristics such as dynamic environment, high-dimensional policy selection, long-term planning requirements, and sparse reward mechanism. 
 Such problems require algorithms to be able to conduct effective policy exploration and long-term planning in an environment where feedback is not provided frequently\cite{bellemare2016intrinsic}. 
 AE-HNN-NCS can effectively pre-select potential optimal policies through HNN, helping the algorithm quickly find solutions when facing multiple enemies and path choices. Its AE can also reduce the dimensionality of high-dimensional policies and reduce computational complexity. 
 In contrast, PLASTIC enhances adaptability to environmental changes by optimizing the plasticity of inputs and labels, but its ability to adjust policies to obtain effective feedback is not as good as AE-HNN-NCS under the sparse reward mechanism. 
 STORM relies on the world model for environmental simulation, and when faced with sparse rewards and complex dynamic environments, it is prone to accumulation of prediction errors, resulting in performance degradation.
 Freeway-like problems have the characteristics of low-dimensional decision space, dynamic environment change, time-sensitive decision-making, and non-sparse reward mechanism\cite{mnih2015dqn}. 
 Due to the relatively simple decision-making process, players mainly rely on quick reactions and accurate judgment to avoid collisions in the traffic flow. Successfully solving such problems requires algorithms to quickly adapt to dynamic environments, implement efficient exploration mechanisms, simplify modeling, and provide fast decision-making and feedback response capabilities. While AE-HNN-NCS has more advantages in high-dimensional, complex problems, it excels in the dynamically changing, low-dimensional environment of Freeway. By combining the strengths of NCS and AE, it maintains efficient policy adjustment in rapidly changing scenarios, allowing the algorithm to respond swiftly to time-sensitive decisions. In contrast, while STORM leverages the sequence modeling capability of the Transformer and can make appropriate decisions in dynamic environments, its complex world model fails to fully capitalize on its strengths in such simpler problems. Similarly, while PLASTIC optimizes the plasticity of inputs and labels, it struggles to fully leverage its flexibility when working with low-dimensional policy spaces.
 %In Pong, a game characterized by precise 
 Pong-like problems are characterized by simple rules and goals, rapid response and real-time decision-making, relative symmetry, and an immediate feedback and reward mechanism. Players need to react quickly in a highly dynamic environment, so algorithms designed to solve such problems must have fast decision-making capabilities, efficient policy search mechanisms, dynamic adaptability, and simple, effective modeling. In AE-HNN-NCS, the AE enables rapid policy adjustment in dynamic environments, ensuring quick decision-making through the HNN. This ability is crucial in handling simple rules and fast-paced problems, as the algorithm can adapt to changes in the game state swiftly and maintain a high win rate. In contrast, STORM  shows lower performance compared to other state-of-the-art algorithms like TWM \cite{robine2023worldmodels} and DreamerV3 \cite{hafner2023wm} in Pong. Its complex world model fails to fully leverage its strengths, instead introducing unnecessary computational overhead in simpler games. AE-HNN-NCS, on the other hand, excels in rapidly changing environments due to its efficient policy search and dynamic adaptability. While PLASTIC improves the plasticity of inputs and labels, it does not achieve the same level of dynamic adaptability as AE-HNN-NCS in quickly changing environments. Although both Pong and Alien are dynamic games, their differences—Pong's simple rules and immediate feedback versus Alien's complex environment and diverse policy requirements—demonstrate that the effectiveness of an algorithm in highly dynamic games depends not only on the dynamics but also on the complexity and strategic demands of the game.
   Regarding our algorithm, the trajectory-like clustering observed in the t-SNE plots reflects the unique mechanism of policy exploration in our approach, as well as the process by which it gradually converges to the optimal policy through multiple explorations and optimizations. This indicates that our model explores the policy space in an orderly manner while progressively optimizing, which helps avoid blind searching. Furthermore, through early extensive exploration, it can identify multiple potential optimal policy paths. This gradual convergence exploration method maintains a certain level of randomness and diversity while ultimately clustering around better solutions. Compared to PLASTIC's rapid adaptation and STORM's random exploration, the AE-HNN-NCS demonstrates a balanced capability for exploration and optimization.

 We also conducted sample analyses on the visualizations of specific Atari games. For details, please refer to AppendixB.

\subsection{How does it perform in continuous action spaces?}
%在连续动作空间中的表现，对policy造成影响
%To comprehensively evaluate the effectiveness of the proposed method, in addition to testing in the discrete Atari environment, we also use the popular continuous Mujoco environment as test problems. 
Testing algorithms in continuous action spaces (e.g., MuJoCo)\cite{todorov2012mujoco} is essential because they fundamentally differ from discrete spaces (e.g., Atari), posing higher demands on policy modeling and optimization. Continuous spaces require policies to output precise continuous values, making exploration and optimization more complex. This testing validates algorithm performance in complex control tasks and evaluates their generality and robustness. These environments are widely used in the RL field~\cite{lillicrap2015ddpg,schulman2017ppo,duan2016benchmark}. These benchmarks are accessible through OpenAI Gym\cite{brockman2016gym}, which provides researchers with an interface for benchmarking their algorithms. The goal of these problems is to apply torques to the robot's joints to complete specific motion tasks.

To evaluate the practical effectiveness of the proposed algorithm, we conducted a comparative analysis with state-of-the-art methods, including TERL\cite{zhu2024twostage}, PROTECTED\cite{liu2024robustrl}, RE$^2$-ERL\cite{hao2022erlre2}, and RL algorithms such as TD3\cite{fujimoto2018td3} and PPO\cite{schulman2017ppo}. The hyperparameters for each algorithm were set to the recommended values from the respective papers or the default values in the authors' implementations. Additionally, we selected four locomotion tasks as comparative instances: Ant, Hopper, Swimmer, and Walker 2D. These four tasks encompass a wide range of motion patterns and control challenges. These tasks were initially introduced as benchmarks in pioneering algorithms like TD3 and PPO and have gradually become standard test sets in reinforcement learning research.

The experimental results are shown in the table. From the table, it can be observed that our algorithm demonstrates optimal performance in Hopper, Swimmer, and Walker2D tasks. Although our algorithm is not the best in the Ant task, it achieves performance close to the optimal level. Ant involves coordinating a four-legged robot to walk on complex terrains, testing an algorithm's high-dimensional control capabilities. The Ant task involves coordinating multiple leg joints, resulting in a large and highly complex policy space. Although our algorithm uses NCS to enhance exploration, it may still exhibit insufficient exploration in high-dimensional tasks, leading to slightly inferior performance compared to RE$^2$-ERL. Hopper is a single-legged jumping robot, suitable for evaluating an algorithm's stability control. Swimmer, a snake-like robot, is used to test an algorithm's optimization capabilities in low-dimensional control tasks. Walker 2D, a bipedal walking robot, can test an algorithm's ability to handle gait planning. 
\begin{table}[tbp]
\caption{Performance results of SOTAs and AE-HNN-NCS on four Mujoco games.}
\label{Tablene twork}
\centering 
\tiny
\begin{tabular}{c |c | c| c| c| c| c}
\toprule
%Problems  & Parameter& Geometry&Modality&Separability\\
\  Games &PPO&TD3&TERL&PROTECTED&RE$^2$-ERL&AE-HNN-NCS\\
\midrule
%Problems  & Parameter& Geometry&Modality&Separability\\
\  Ant &2937&6690&6720&5863&\hl{6947}&6843\\
\  Hopper &1892&3447&3711&3658&3229&\hl{3797}\\
\  Swimmer &337&361&360&342&354&\hl{363}\\
\  Walk2d &3441&5583&6158&6309&6182&\hl{6324}\\
\bottomrule
\end{tabular} 
\end{table}

\section{Conclusion and Future Work} \label{sectionConclusion}
% \textcolor{black}{Background and challenges\\
% our method\\
% specifically, modules/components of the algo\\
% results\\
% impact of this work on the research topic\\
% limitation of the algorithm\\
% future work
% complex problem application, algorithm upgrading\\
% }
This paper presents the AE-HNN-NCS algorithm, designed to address the challenges of the curse of dimensionality and low search efficiency in deep reinforcement learning. The algorithm incorporates an AE and a HNN. Building on the previous work of the PE-SAERL method \cite{tang2022pe}, our proposed algorithm enhances the original policy embedding module. The new AE module not only reduces dimensionality but also extracts key features, enabling more effective pre-selection of policies. Additionally, by employing HNN as the agent module, the algorithm benefits from improved learning capabilities and more accurate classification. Experimental results demonstrate that AE-HNN-NCS performs exceptionally well across 10 Atari games, outperforming most baseline and state-of-the-art methods.

Our work demonstrates that addressing the curse of dimensionality requires not only dimensionality reduction but also consideration of important features in the data to fully exploit useful information. Although the techniques used in our approach are simple, we believe they can provide valuable insights for future research. However, several aspects need further exploration. First, we used basic autoencoders, whose feature extraction capabilities are still limited. Future research could investigate the use of more advanced autoencoder variants, such as variational autoencoders or convolutional autoencoders, as well as other dimensionality reduction techniques. Second, in our approach, HNN is used as a classification model to select the candidate policies with the highest probability as the best offspring. An interesting direction for future work would be to modify HNN into a regression model to predict the scores of candidate policies, which could potentially improve the accuracy of pre-selection. 

Moreover, regarding result diversification, Qian et al. \cite{qian2022diversification} suggested that multi-objective evolutionary algorithms could be an effective approach. Future research could draw on their theoretical framework to enhance our algorithm's ability to generate a more diverse set of candidate policies, which may improve the overall exploration capacity of evolutionary reinforcement learning in complex environments. 

Finally, our experiments were limited to the Atari and Mujoco environments. Future work should test the algorithm's performance in more complex real-world scenarios and other simulated environments, such as cloud computing and the traveling salesman problem \cite{yang2024reducing,hong2020efficient,hong2019scalablemoea,liu2023humanassist}. Additionally, a theoretical analysis of the proposed method could offer deeper insights \cite{qian2013recombination,bian2025stochastic}.
\bibliographystyle{ACM-Reference-Format}
\bibliography{tevcref}
\end{CJK}
\end{document}